\documentclass[pdflatex,sn-mathphys]{sn-jnl}

\usepackage{graphicx}
\usepackage{multirow}   
\usepackage{amsmath}
\usepackage{color}

\jyear{2021}%

\theoremstyle{thmstyleone}%
\theoremstyle{thmstyletwo}%

\theoremstyle{thmstylethree}%

\begin{document}

\title[Relational Chain Reasoning Improved Embedded KGQA]{Improving Embedded Knowledge Graph Multi-hop Question Answering by Introducing Relational Chain Reasoning}

\author[1]{\fnm{Weiqiang} \sur{Jin}}\email{weiqiangjin@stu.xjtu.edu.cn}

\author[1]{\fnm{Biao} \sur{Zhao}}\email{biaozhao@xjtu.edu.cn}

\author*[2]{\fnm{Hang} \sur{Yu}}\email{yuhang@shu.edu.cn}

\author[2]{\fnm{Xi} \sur{Tao}}\email{20721546@shu.edu.cn}

\author[3,4]{\fnm{Ruiping} \sur{Yin}}\email{yinruiping@bjut.edu.cn}

\author[1]{\fnm{Guizhong} \sur{Liu}}\email{liugz@xjtu.edu.cn}

\affil[1]{\orgdiv{School of Information and Communications Engineering}, \orgname{Xi`an Jiaotong University}, \orgaddress{
\city{Xi`an}, \postcode{710049}, \state{Shaanxi}, \country{China}}}

\affil[2]{\orgdiv{School of Computer Engineering and Science}, \orgname{Shanghai University}, \orgaddress{
\city{Baoshan}, \postcode{200444}, \state{Shanghai}, \country{China}}}


\affil[3]{\orgdiv{Information Faculty of Computer School}, \orgname{Beijing University of Technology}, \orgaddress{
\city{Chaoyang}, \postcode{100124}, \state{Beijing}, \country{China}}}

\affil[4]{\orgdiv{Engineering Research Center of Intelligence Perception and Autonomous Control}, \orgname{Ministry of Education}, \orgaddress{
\postcode{100124}, \state{Beijing}, \country{China}}}


\abstract{
Knowledge Graph Question Answering (KGQA) aims to answer user-questions from a knowledge graph (KG) by identifying the reasoning relations between topic entity and answer. As a complex branch task of KGQA, multi-hop KGQA requires reasoning over the multi-hop relational chain preserved in KG to arrive at the right answer. Despite recent successes, the existing works on answering multi-hop complex questions still face the following challenges: i) The absence of an explicit relational chain order reflected in user-question stems from a misunderstanding of a user's intentions. ii) Incorrectly capturing relational types on weak supervision of which dataset lacks intermediate reasoning chain annotations due to expensive labeling cost. iii) Failing to consider implicit relations between the topic entity and the answer implied in structured KG because of limited neighborhoods size constraint in subgraph retrieval-based algorithms. 
To address these issues in multi-hop KGQA, we propose a novel model herein, namely Relational Chain based Embedded KGQA (Rce-KGQA), which simultaneously utilizes the explicit relational chain revealed in natural language question and the implicit relational chain stored in structured KG. Our extensive empirical study on three open-domain benchmarks proves that our method significantly outperforms the state-of-the-art counterparts like GraftNet, PullNet and EmbedKGQA. Comprehensive ablation experiments also verify the effectiveness of our method on the multi-hop KGQA task. We have made our model's source code available at github: \url{https://github.com/albert-jin/Rce-KGQA}.}

\keywords{Data Mining and Search, Question Answering, Knowledge Graph based Multi-hop QA, Neural Semantic Parsing, Knowledge Graph Embedding}



\maketitle

\section{Introduction}
\label{introduction}
Knowledge Base Question Answering (KBQA) \cite{xiaojin2019chatrobot} is an attractive service mining and analytics method that has attracted extensive attention from academic and industrial circles in recent years. Given a natural language question, the KBQA system aims to answer the correct target entities from a given knowledge base (KB) \cite{fu2020surveyQA}. It relies on certain capabilities including capturing rich semantic information to understand natural language questions clearly and seek correct answers in large scale structured knowledge databases accurately.  Knowledge Graph Question Answering (KGQA) \cite{hao2017endQA,michael2018SimpleQ} is a popular research branch of KBQA which uses a knowledge graph (KG) as its knowledge source \cite{fu2020surveyQA,yunshilan2021surveyQA} and uses factoid triples stored in KG to answer natural language questions. Thanks to KG's unique data structure and its efficient querying capability, users can benefit from a more efficient acquisition of the substantial and valuable KG knowledge, and gain excellent customer experience.

Early works \cite{yunshilan2019aQA,yunshilan2019bQA} on KGQA focus on answering a simple question, where only a single relation between the topic entity and the answer are involved. For example, in the question ``What films did [Martin Lawrence] act in?'', as depicted in Fig.\ref{graph_example}, there only exists a single relation `starred\_actors' between the topic entity `Martin Lawrence' and the answer. The final answer only relies on just a single KG fact (Martin Lawrence, starred\_actors\_reverse, Black Knight). To solve simple question tasks, most traditional methods \cite{bast2015fbQA,abujabal2017atgKGQA} create diverse pre-defined manual templates and then utilize these templates to map unstructured questions into structured logical forms. Unfortunately, these pre-defined templates and hand-crafted syntactic rules are both labor-intensive and expensive. Moreover, such approaches require crowd workers to be familiar with linguistic and specific domain expert knowledge. Due to the dependency on large-scale fixed rules and manual templates, these methods cannot handle complex questions which require multiple relations inferences.

To make KGQA more applicable in realistic application scenarios, researchers have shifted their attentions from simple questions to complex ones 
. Knowledge graph multi-hop question answering is a challenging task which aims to seek answers which is multiple hops away from the topic entity in the knowledge graph. For example, the question ``Who directed the films which [Martin Lawrence] acted in?'' is a complex multi-hop question which requires a relational chain (starred\_actors, directed\_by) which has multiple relationships to arrive at the corresponding answers. This task is a relatively complex task compared with its simple counterparts due to the multi-hop relations retrieval procedure \cite{yunshilan2021surveyQA} which requires more than one KG fact in the inference.

Previous approaches \cite{pullnet,srn,dong2015CnnQA,yuhang_svm} for handling complex question answering constructed a specialized pipeline consisting of multiple semantic parsing and knowledge graph answer-retrieving modules to complete the KGQA task. However, most have several drawbacks and encounter various challenges. We summarize these challenges as follows:
 
 \begin{figure}[h]
\centering
\includegraphics[width=1.0\linewidth]{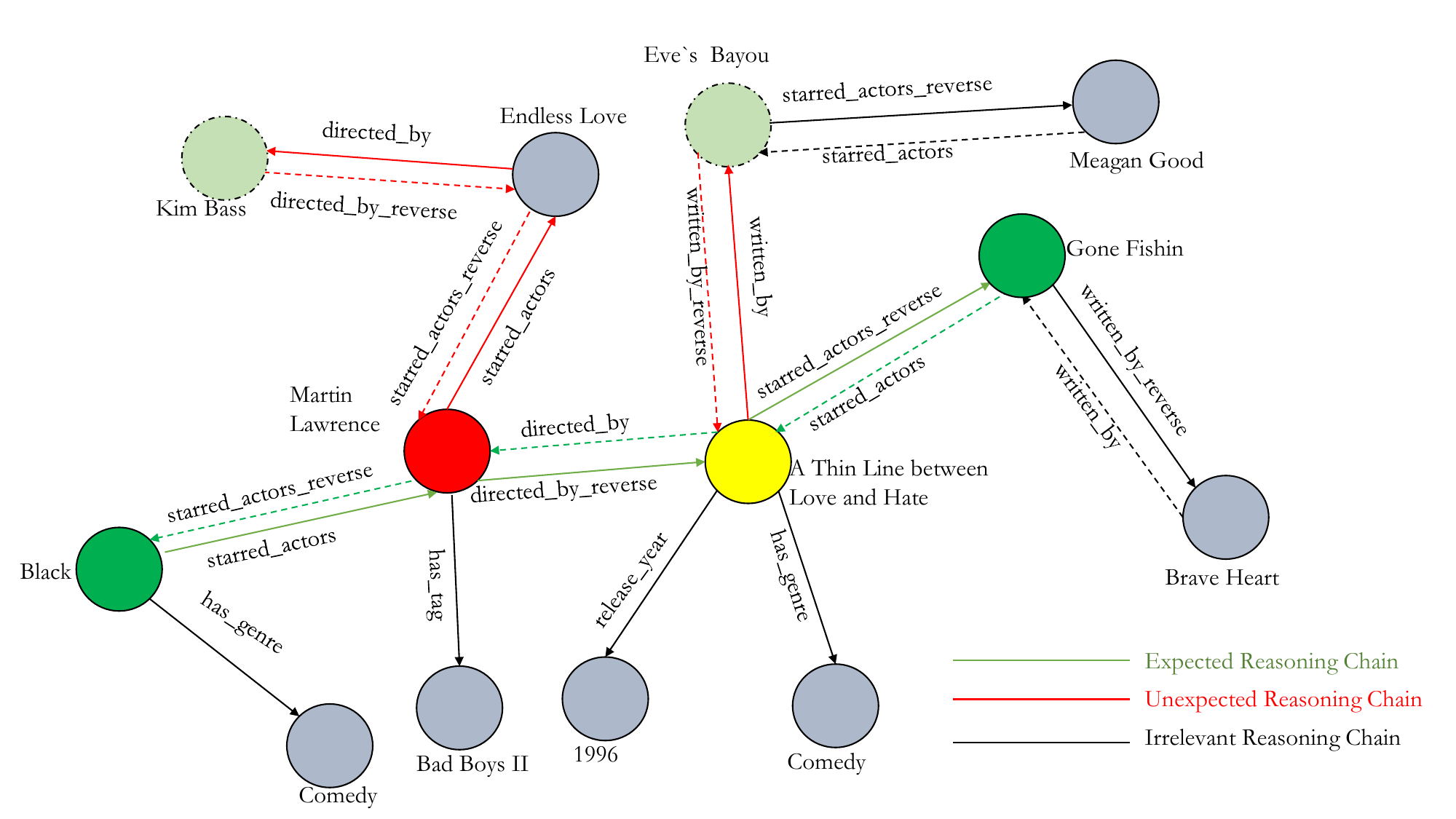}
\caption{The Freebase subgraph entered on the topic entity [Martin Lawrence] of the example questions. The red, yellow, green, green with an imaginary line, and grey circles denote the topic, intermediate, expected, unexpected and irrelevant entity nodes, respectively. The green, red and grey colored edges indicate the correct, incorrect and irrelevant reasoning chains, respectively.}
\label{graph_example}
\end{figure}
 
\textbf{Unexpected Relation Type Recognition.}
 As the first step of KGQA, a semantic parser of most existing methods performs with poor accuracy in recognizing the correct relational type implied in questions, which hinders downstream answering reasoning. For example, as shown in Fig.\ref{graph_example}, let us consider questions where the topic entity and answer are connected by a multiple-hop reasoning chain, e.g., ``Who acted in the movies directed by the director [Martin Lawrence]?''. To answer this type of question, the related two facts (Martin Lawrence, directed\_by\_reverse) and (A Tine Line, starred\_actors\_reverse, Gone Fishin) help derive the answers within the neighborhood of the topic entity [Martin Lawrence]. Typically, these methods are prone to encounter incorrect relational reasoning ($AB \rightarrow AC$) when we mistake the unexpected relational chain (starred\_actors\_reverse, written\_by) for the expected relational chain (starred\_actors\_reverse, directed\_by\_reverse). Thus, it is necessary to optimize relational semantics parsing for more accurate user intention recognition.
 
\textbf{Unexpected Relation Order Recognition.}
Semantic parsing-based \cite{tois1,tois3,yuhang_csvm,xianan2022machinelearning} methods mostly do not effectively capitalize on the correlation information of relationship order and direction from user-question expression. They become more susceptible to incorrect understanding when the questions are complicated from both semantic and syntactic aspects. The accuracy rate of parsing syntactics can be dramatically decreased by those with long-distance dependency. More especially, tracing back to the above question example and Fig.\ref{graph_example}, in addition to the correct chain (with green arrows), the spurious multi-hop chain (with red arrows) from entity node [Martin Lawrence] to [Kim Bass] can lead to incorrect reasoning results when the semantic parser module fails to parse such semantics as, reversing ($AB \rightarrow BA$) or shuffling ($ABC\rightarrow BCA$), the correct order of the relational chain. In short, we need an accurately capture of longer ordered-relational mappings implied in user language expressions to reach correct answers.

\begin{figure}[h]
\centering
\includegraphics[width=0.95\linewidth]{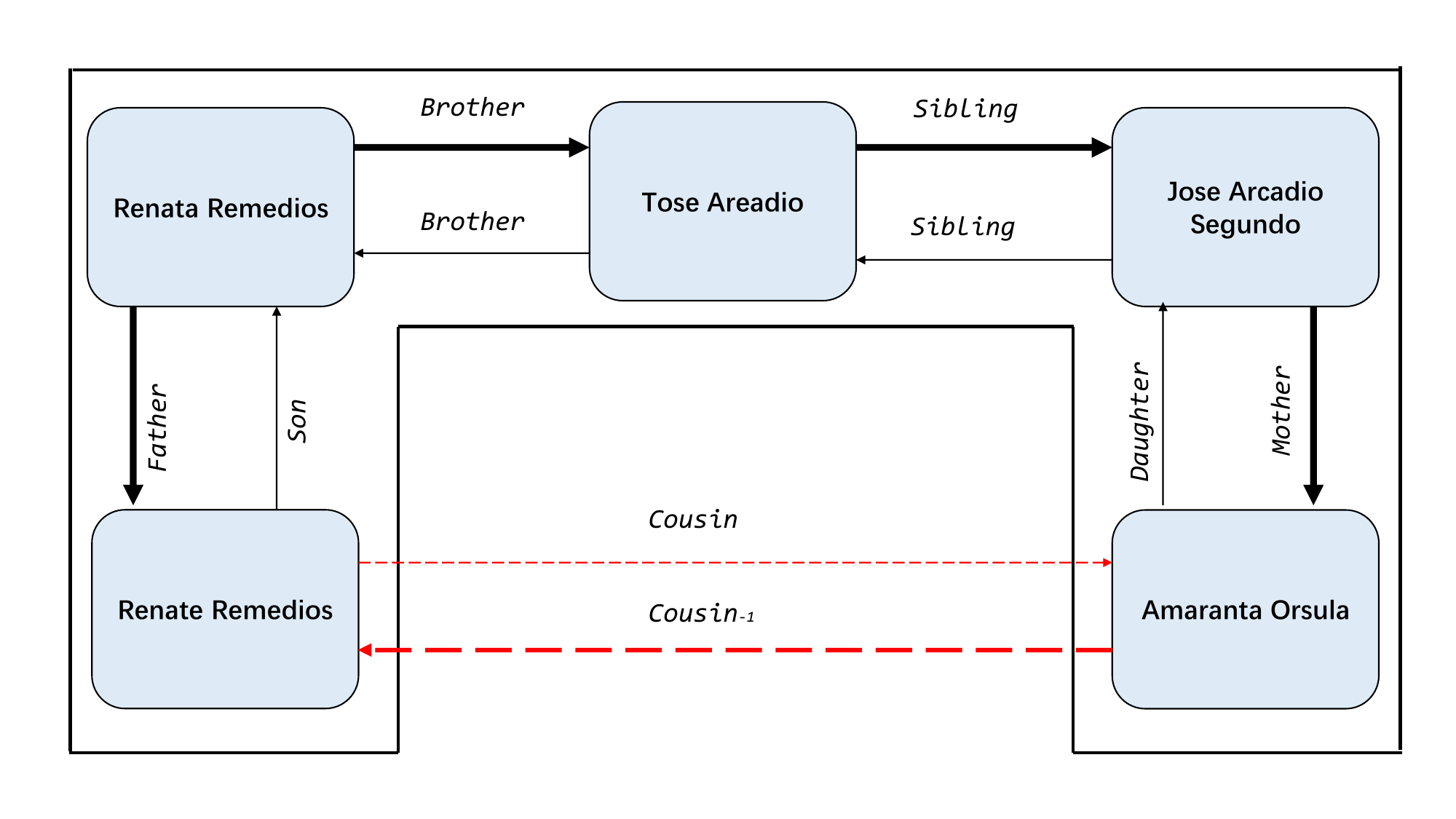}
\caption{A structured knowledge graph related to the user-question, ``Who is [Renate Remedio]'s cousin?''. The illustration typically introduces implicit relations between the topic entity and the answer that hides in KG.}
\label{question_example}
\end{figure}

\textbf{Implicit Relation Reasoning \& Subgraph Neighborhood Constraint.}
Most mainstream KGQA methods \cite{tois2,tois4} cannot indirectly capture knowledge of implicit relational chains for reasoning due to the limiting constraint of neighborhood size constraint. All the answers are provided by retrieving the extracted question subgraph. Let us consider the question ``Who is [Renate Remedio]'s cousin?''. As depicted in Fig.\ref{question_example}, the corresponding knowledge graph has no direct relational chain between the topic entity [Renate Remedio] and answer [Amaranta Orsule]. In other words, for future successive KGQA solutions, it is important to be able to discover the implicit factoid knowledge (Remedio $\leftarrow$ Cousin $\rightarrow$ Orsule) in the incomplete KG by the explicit relational chain (Remedio $\leftarrow$ Father - Siblings - Mother $\rightarrow$ Orsule), similar to the KG-based link-prediction task \cite{Nathani2019gat}. Furthermore, most existing methods labor under the undesirable constraint of answer detection from a pre-specified localized KG sub-graph neighborhood. For example, the state-of-the-art (SOTA) method GraftNet \cite{sun2018graftnet} whose answer is restricted to being a subset of the entities present in a localized KG sub-graph neighborhood, only reports a recall around 0.55 on an incomplete KG where only half of the original triples are presented.

To alleviate these limitations and challenges for the multi-hop KGQA task, our paper introduces a novel architecture, namely Relational Chain based Embedded KGQA, which supports the integration to learn the explicit relational mappings implied in the user's expression and the implicit knowledge from the structured KG, similar to link prediction. Our proposed approach counters these limitations by simultaneously using the explicit semantic relational chain described in the question and the implicit relational chain between the structured KG nodes. We use the knowledge graph embedding and construct the \textit{Answer Filtering Module} to calculate the mutual relationship between the topic entity and answer.
Motivated by the previous work EmbedKGQA \cite{apoorv2020embedKGQA}, we show how our model leverages an end-to-end neural network that employs the KG entity and relation embeddings to provide complex questions with answers from the KG. Since our model replaces the traditional pipeline procedure of generating and retrieving a localized subgraph at intermediate reasoning steps, it helps to decrease memory costs efficiently and obtain computational efficiency. To obtain  a more competitive performance in large-scale KG, we apply an extra reasoning procedure called the \textit{Relational Chain Reasoning Module} to prune the candidate entities ranked by the \textit{Answer Filtering Module}. We apply a Siamese architecture \cite{sia2016jonas} based on the \textit{long short-term memory } (LSTM) \cite{gjq2022kbs} and transformer \textit{RoBERTa} \cite{roberta} to learn the semantic similarity between the relational chain of the problem description and the KG factoid relational chain. It also leverages the external supervised signal of the relational chain from the training sample. By calculating the semantic similarity between question semantics and candidate entity retrieval chain, we can further determine the final answer more accurately. The internal construction details of our model are introduced in Sec. \ref{modeldetails}.

We summarize the contributions of this paper as follows:

1. We propose a novel approach namely Relational Chain based Embedded KGQA which includes two main modules: the \textit{Answer Filtering Module} and the \textit{Relational Chain Reasoning Module}. Moving away from previous studies, our model simultaneously takes advantage of the knowledge graph embedding and training with weak supervision by predicting the intermediate relational chain implied in KG to perform the multi-hop KGQA task.

2. We introduce the \textit{Answer Filtering Module}, a knowledge graph embedding based end-to-end network for preliminary answer filtering. This module can address the problem of inadequacy that is a factor in the missing links in the incomplete knowledge graph thanks to its capability of capturing implicit KG relationships. Furthermore, we consider all entities as candidate answers in this step, so our model won't suffer from the out-of-reach issues brought by limited subgraph neighborhood constraint.

3. Our proposed \textit{Relational Chain Reasoning Module} can help capture the multi-hop relations surrounding the topic node to support the results more accurately. We apply the Siamese network \cite{sia2016jonas} to calculate the vector representation-based semantic similarity score between the user's question and KG structured knowledge. To the best of our knowledge, our proposed sub-module is the first to consider the question relational direction and order information by using the Siamese network.

4. Our experimental results on three widely adopted KGQA benchmarks demonstrate our method's competitive capability compared with most SOTA methods (average 1.2\% absolute improvement across $hit@1$ evaluation metric). Furthermore, using an extensive ablation study, we demonstrate the superiority and effectiveness of our proposed model for the multi-hop KGQA task.

The rest of the paper is organized as follows: We first provide a thorough review of the related KGQA works in Sec. \ref{relatedwork}. Next, we introduce the preliminary knowledge about the KGQA task in Sec. \ref{preliminary}. Following the internal structure of our model, we then explicate our two features: \textit{Answer Filtering Module} in Sec. \ref{entitydetails} and \textit{Relational Chain Reasoning Module} in Sec. \ref{relationdetails}. Sec. \ref{experiment} describes the experimental details on three open-domain datasets. Finally, in Sec. \ref{conclusion}, we conclude our contributions to this work and suggest several promising innovations for our Relational Chain based Embedded KGQA in the future.

\section{Related Work}\label{relatedwork}

Our work is closely related to the Multi-hop Knowledge Graph Question Answering, Knowledge Graph  Embedding, Siamese Network, and Pretrained Language Model.
\subsection{Multi-hop KGQA}
\textit{Multi-hop \textbf{K}nowledge \textbf{B}ase \textbf{Q}uestion \textbf{A}nswering} comprises two mainstream branches: Information Retrieval-based (IR) and Semantic Parsing-based (SP-based). The most popular methods fall into these two categories.

\subsubsection{Semantic Parsing Methods} \textbf{SP}-based approaches follow a parse-then-execute procedure. These methods \cite{yunshi2020graphKGQA,he2021stu-teaQA,huang2019KgeQA,apoorv2020embedKGQA,chen2021foodQA} can be summarized as the following steps: (1) \textit{question semantic understanding}: parsing relations and subjects involved in complex questions, (2) \textit{logical formula construction}: decode the subgraph into an executable logic form such as high-level programming languages or structured queries such as SPARQL, (3) \textit{KG-based positioning and querying}: search from KGs using the query language and provide query results as the final answer.
Owing to its intermediate procedure of generating expressive logic forms, SP-based methods are more interpretable than their IR-based methods counterparts. Nonetheless, for most existing SP-based methods, more relations in complex questions indicate a larger search space of potential logic forms for parsing, which will dramatically increase the computational cost. 

Yu et al. \cite{improvedRDKGQA} pointed out that KG relation detection is a core component and entity linking is a key step in KGQA tasks. To improve the recognition accuracy of both sub-tasks, they proposed the Hierarchical Residual BiLSTM (HR-BiLSTM) to encode question descriptions and word-level and phrase-level relationship path. The new HR-BiLSTM module calculates the similarity scores for all the questions and textual relationships, which integrates for these two components entity linking and relationship path identification into a single step and enhances each other. When in inference, the model only selects the highly-scored (relations, topic entity) pairs as correct answers from candidates.


Miller et al. \cite{kvmem} proposed an ideal domain-specific KGQA framework, called Key Value-Memory networks (KV-MemNN), which has proved to be effective to support answer reasoning over specific domain multi-source knowledge like textual documents and structured KG. It performs QA by employing a widely used long-term memory mechanism to reason on a key-value structured memory network. 
They defined three operations, i.e., key hashing, the model first fetchs all KG triples relevant to given questions and then stores their topic entities and relationships in the key slot, tail entity in the value slot; key addressing, the model assigns each memory unit with a normalized relevance weight by the dot product operation as the relevance probability between the question and each key representations in the memory; finally, value reading, where the model reads the values of all addressed memories by taking their weighted sums of all values and relevance weights, and use the outputs to represent intermediate reasoning results, which is then used to update the question representation. To obtain the final prediction over all candidate answers, the model repeats the key addressing and value reading steps in the Ranking component several times. 

However, the KV-MemNN obviously presents the following challenges: 1) It often fails to precisely update multi-relation question queries during multiple memory reading. 2) It reads the memory repeatedly since they can not well determine accurately when to stop. 3) It focuses more on memory facts understanding rather than the properly questions understanding, so it does not perform as well as expected when applied to the scenario where its questions are complicated and associated with complex constraints, such as an open-domain KGQA task. 4) It selects the candidate with the highest similarity score as the only answer in default. So, it conducts inefficiently when the questions contain more than one answer.

To solve these challenges, Xu et al. \cite{enhancedkvmem} proposed an interpretable mechanism to enable a basic KV-MemNN model to work for complex questions, which yielded state-of-the-art performances on three benchmarks. Enhanced KV-MemNN introduced a novel \textbf{STOP} strategy into multi-hop memory reading to generate a flexible number of queries and introduce a new query updating method, which considers the already-addressed keys in previous hops as well as the value representations that avoids repeated or invalid memory readings. For multi-constraint questions, the model considers the value representation of each hop by accumulating all the value representations of both current and previous hops to address each relevant constraint at different hops.

In addition to the above representative methods, many knowledge base question answering approaches based on Graph Neural Network (GNN) \cite{linetal} and Graph Convolutional Network (GCN) \cite{QCNpaper} have been proposed in recent years, approaches such as Graph Convolutional Network-based Multi-Relation Question Answering system (QAGCN) \cite{wangQAGCN} and Case-Based Reasoning SUBGraph model (CBR-SUBG) \cite{pmlrv162das22a}. As the name suggests, QAGCN is a simple but effective model that leverages attentional graph convolutional networks that can perform multi-step reasoning during the encoding of knowledge graphs. Able to leverage highly-efficient embedding computations, the model's significant advantage is that it can essentially simplify complex reasoning mechanisms. CBR-SUBG is a semiparametric model for weakly-supervised KGQA that retrieves similar queries and utilizes the similarities in graph structure of local subgraphs to answer a query. It contains a parametric component comprising a graph neural network (GNN). Through experiments, it performs competitively with state-of-the-art KGQA models on multiple benchmarks. Due to its capacity for reasoning pattern identification, the method CBR-SUBG can also provide interpretable paths for returned answers, which could bring slightly better interpretability.


\subsubsection{Information Retrieval Methods} \textbf{IR}-based approaches typically include a series of procedures as follows: question-specific graph extraction, question semantics representation, extracted graph-based reasoning and answer candidates ranking. Given a complex question description, these methods \cite{chen2019uhopKGQA,jain2016fmnKGQA} first construct a question-specific subgraph which includes all question-related entity nodes and relation edges from KGs without generating an executable logic formula, This is followed by employing a question representation module to encode user-question tokens as low-dimensional vectors. Secondly, an extracted-graph based reasoning module conducts a semantic matching algorithm to aggregate the center entity's neighborhoods' information from the question-specific subgraph. At the end of the reasoning, they rank all the entities' scores in the subgraph by applying an answer-ranking module to predict the top-ranked entities as the final answers. Based on feature representation technology, IR-based approaches can be divided into feature engineering-based approaches and representation learning-based approaches. 


\textit{IR-based} feature engineering approaches \cite{irfrmethod} rely on manually defined and extracted features, which are time-consuming and cannot detect the whole question semantics. To solve these problems, representation learning \textit{IR-based} methods convert questions and related entities into distributed vector representations in the same dimension space and treat KGQA tasks as semantic matching between distributed representations of questions and candidate answers \cite{fu2020surveyQA}. 

Sun et al. \cite{sun2018graftnet} propose a integrated framework namely \textit{GRAFT-Net}, which is adopted an knowledge fusion strategy, where the answers are selected from a heterogeneous question-specific subgraph constructed from the KG and textual documents based on the given questions. The subgraph contains three factors: entity nodes, sentence nodes and a special type of edges which indicates the mutual relations between entity and sentence nodes. During answer detection, the convolution neural network GRAFT-Net spreads central entity node feature to neighboring nodes in several iterations and determines whether an entity node is an answer or not.

However, the automatically constructed subgraph in GRAFT-Net relies heavily on heuristic rules and can lead to serious error cascading and bring incorrect reasoning. Thus, soon after proposing the GRAFT-Net \cite{sun2018graftnet}, Sun et al. \cite{pullnet} presented a learned iterative process for topic-entity-centric graph construction. The improved method, called \textit{Pull-Net}, where the ``pull'' classifier is weakly supervised so that only QA pairs are used for supervision. It first selects seed entity nodes by GRAFT-Net and a novel classification model at each step. Then, more and more extra valuable entities and sentences are introduced into the current graph through several pre-defined operational iterations, with the final answer determined by the same procedure as GRAFT-Net \cite{sun2018graftnet}. Experimentally, PullNet improves dramatically over prior state-of-the-art methods \cite{sun2018graftnet,jain2016fmnKGQA,irfrmethod} even under weakly supervised signals and incomplete KGs.

A significant challenge in multi-hop Knowledge Base Question Answering is the lack of supervision signals at intermediate steps. To address this challenge, He et al. \cite{he2021stu-teaQA} propose an elaborate teacher-student framework by adapting the generic Neural State Machine (NSM) \cite{nsm} as the student network, while the teacher network aims to learn intermediate supervision signals to improve the student network. The extensive evaluation results with three benchmark datasets show that their proposed model is superior to previous methods in terms of effectiveness for the multi-hop KGQA task. Moreover, other detailed experiments prove that their approach is more flexible to extend itself to other neural architectures or learning strategies on graphs.


Apart from these traditional subgraph-generation methods, researchers also try to incorporate the KG embedding mechanism as extra information into entity and relation representations to alleviate the incomplete KG sparsity problems. Inspired by relationship completion and missing link prediction tasks in the KGs, Saxena et al. \cite{apoorv2020embedKGQA} propose a novel framework, named \textit{EmbedKGQA}, which leverages the pre-trained KG embeddings to enrich the learned entity and relation representations. Extensive comparative experiments on multiple benchmarks show that EmbedKGQA is particularly effective in performing multi-hop KGQA over sparse KGs.

\subsection{Knowledge Graph Embedding}
\textit{Knowledge Graph Embedding} \cite{huang2019KgeQA} is to embed a KG's factoid triples knowledge including all entities and relations into continuous and low-dimensional embedding representation space, such that the original entities and relations are well preserved in the vectors. Representative KG embedding models exploit the distance-based scoring function $f(\cdot)$ which is used to measure the plausibility of a triple \textit{(topic entity, predicate, and tail entity)} as the distance between the head and tail entity such as TransE \cite{bordes2013transe} and its extensions (e.g. TransH \cite{zwang2014transh}), DistMult \cite{yang2015distmult} and ComplEx \cite{trouillon2016ComplEx}. In short, a typical KG embedding technique generally consists of three steps: (1) representing entities and relations, (2) defining a scoring function, and (3) learning entity and relation representations. Thanks to its ability to simplify the manipulation while preserving the KG inherent structure, it can benefit a variety of downstream tasks to take the entire KG into consideration, such as entity alignment \cite{sunwang2020align}, relation prediction \cite{Nathani2019gat} and even KGQA work \cite{apoorv2020embedKGQA}. The effectiveness of knowledge graph  embedding in various real-world NLP tasks \cite{jiangwanmeies,alies} motivates us to explore its potential advantages in the KGQA task.

\subsection{Siamese Network}
The \textit{Siamese network} \cite{sia2016jonas} is a semantic textual similarity metric that is built on top of a feature representation network such as CNN \cite{yanlecun2005cnnsia} and RNN \cite{sia2016jonas}. Given an example input pair, the Siamese network first maps these two inputs into sequences of the word embedding vector using large-scale pretrained embeddings like Glove \cite{jeffrey2014glove}, then passes these vectors through the representation extractor's forward procedure and get the semantic vector representations, respectively. Finally, the Siamese network applies $\ell^{1}$ norm (Manhattan distance) or $\ell^{2}$ (Euclidean distance) norm as the distance measurement function to calculate the similarity between these two representations. Furthermore, the long short-term memory (LSTM) is superior to the original RNNs for learning long-range dependencies because its memory cell units can capture rich features across lengthy language token sequences. Therefore, in this work, we employ the Siamese adaptation of the bidirectional LSTM network to learn the semantic relational chain. 

\subsection{Pretrained Language Model: RoBERTa}
Bidirectional Encoder Representations from Transformers \cite{google2018bert}, or BERT, is a revolutionary self-supervised pretraining technique that learns to predict intentionally hidden (masked) sections of text. Crucially, the representations learned by BERT have been shown to generalize well to downstream tasks and, when BERT was first released in 2018, it achieved state-of-the-art results on many NLP benchmark datasets.

RoBERTa \cite{roberta}, as is proposed by Liu et al., is built on BERT’s language-masking strategy and modifies key hyperparameters in BERT, and it can be regarded as a heavily optimized version of BERT. It includes removing BERT’s next-sentence pretraining objective, and trains with much larger mini-batches and learning rates. It was also trained on an order of magnitude more data than BERT, for a longer amount of time. This allows RoBERTa representations to generalize even better to downstream tasks compared to BERT. The improved model RoBERTa achieves state-of-the-art results on GLUE, RACE and SQuAD benchmarks, without multi-task finetuning for GLUE or additional data for SQuAD.

\section{Preliminaries}
\label{preliminary}
In this section, we formally introduce the preliminary knowledge \cite{yunshilan2021surveyQA,fu2020surveyQA} on the multi-hop KGQA task formulation and its related definitions. Before the formulaic description, all the summarized pre-defined notations for the KGQA task are given as follows: we denote a KG as $\mathcal{G}(\varepsilon, \mathcal{R})$ in which $\varepsilon$, $\mathcal{R}$ respectively denote the entities and relation set, and we use $(h, \ell, t)$ to represent a factoid triple in KG. We use an uppercase and lowercase letter to denote a matrix (e.g. \textbf{W}) and a vector (e.g. \textbf{v}). The $\ell^{n}$ norm of a vector is denoted as $\|\mathrm{p}\|_{n}$.

\textbf{Definition 1 (Multi-hop question)} \cite{apoorv2020embedKGQA,fu2020surveyQA,yunshilan2021surveyQA} If a natural language question involves more than one predicate between the topic entity and answer, then we believe the answer is multiple hops away from the topic entity in the KG. Thus, we identify this as a multi-hop question. For example, let us consider the multi-hop question: ``When did the film production company announce which actor also directed the movie [Cast a Deadly Spell]?'', which consists of several predicates which correspond to the KG relational links: \textit{release\_year, starred\_actors, directed\_by} respectively.

\textbf{Definition 2 (Knowledge Graph  Embedding)} \cite{quan2017surveyofKGE} The KG embedding algorithm \cite{huang2019KgeQA,quan2017surveyofKGE} aims to map all the KG components including entity and relation to a low-dimension and continuous vector space. Given a KG consisting of $n$ entities and $m$ relations, we firstly initialize the values of $h$, $\ell$ and $t$ randomly. Then, a scoring function $f_{\ell}(h, t)$ which we defined measures the relation of a fact triple $(h, \ell, t)$. Finally, the embedding algorithm utilizes a margin-based ranking criterion to optimize the embedding distribution that maximizes the overall plausibility of factoid triples $(h, \ell, t)$ and to minimize the plausibility of spurious triples $(h^{\prime}, \ell^{\prime}, t^{\prime})$ simultaneously.

\textbf{Definition 3 (\textit{Multi-hop} KGQA task)} \cite{fu2020surveyQA,yunshilan2021surveyQA}
The multi-hop question was introduced in Definition 1. In this section, we define a knowledge graph (KG) as $\mathcal{G}$. $\mathcal{G}$ is a directed graph whose nodes represent entities and edges represent relations, and each triple in the KG represents an atomic realistic fact, such as (Joseph Robinette Biden, president\_of, USA).

Formally, given a complex natural language question in the format of a sequence of tokens $\mathcal{q} = {w1,w2, ...,wl}$ and the available KG $\mathcal{G}$, the KGQA task first links the topic entity ${wi, ...,wj}$ to the KG $\mathcal{G}$. The subject mentioned in a question is also named as a  topic entity. Then, it identifies the most possible KG relations which are related to the user's question. Using these two steps, the goal of KGQA is to determine the factual answer with triples stored in KG, denoted by the set $\mathcal{A}_{q}$, to query $q$ from the candidate entities $E$ by leveraging the topic entity and related relations in KG. Specifically, we focus on solving complex question answering, termed the \textit{multi-hop} KGQA task, where the answer is multiple hops away from the topic entity in a knowledge graph, which means these questions require more than one KG triple.

\begin{figure}[ht]
\centering
\includegraphics[width=0.95\linewidth]{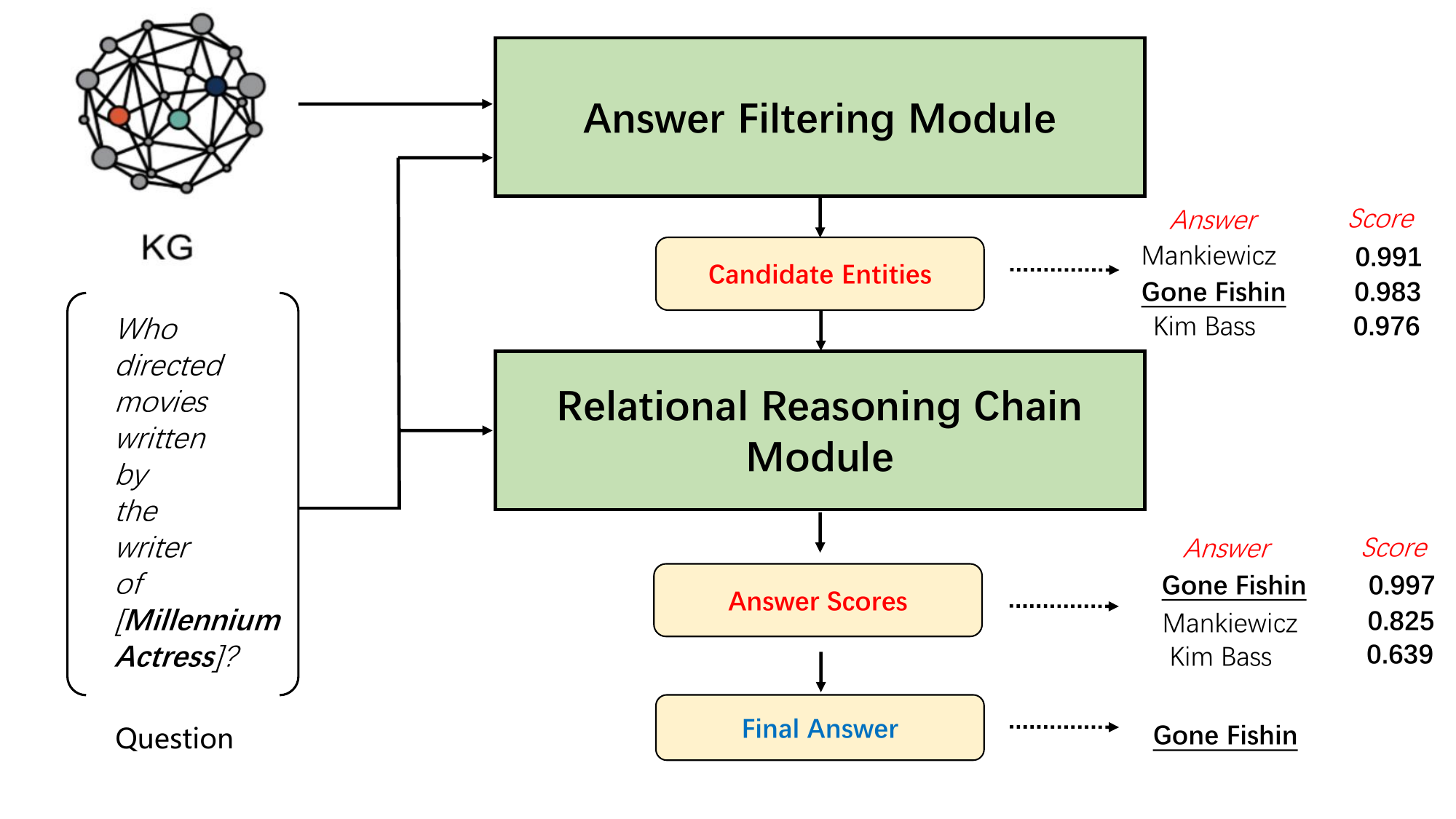}
\caption{This figure shows our overall pipeline architecture for the multi-hop KGQA task. The green rectangles denote our two sub-modules, the solid arrows and dashed arrows indicate the information flowing through our model and the intermediate results. The next figures~[\ref{answer_filtering}, \ref{relational_chain_reasoning}] also illustrate this by using the typical user's complex question ``Who acted in the movies directed by the director [Martin Lawrence]?''}
\label{all_architecture}
\end{figure}

\section{Our Proposed Model}\label{modeldetails}
Our \textit{Relational Chain based Embedded KGQA} is a two-stage pipeline model which consists of two components: \textit{Answer Filtering Module} and \textit{Relational Chain Reasoning Module}.

\subsection{Overview}
\label{overview}
As illustrated in Fig.\ref{all_architecture}, given a real-world question and an available KG, the \textit{Answer Filtering Module} first jointly leverages topic entity embedding and question representation to score all possible candidate entities in this KG to provide a set of pruned candidate answers for this question. However, the entity nodes in a KG are often on a scale as large as a million, hence it could be noisy and inaccurate when comparing the topic entity with all other entities $\hat{\mathbf{t}}$. To make the learning more efficient and accurate, we do not directly select the top-1 scoring entity from the sorted entities as our final answer. Instead, we introduce the extra module \textit{Relational Chain Reasoning Module} to take the relation type and order of semantic relational chain into consideration for a higher $hit@1$ accuracy result.

\begin{figure}[h]
\centering
\includegraphics[width=1.02\linewidth]{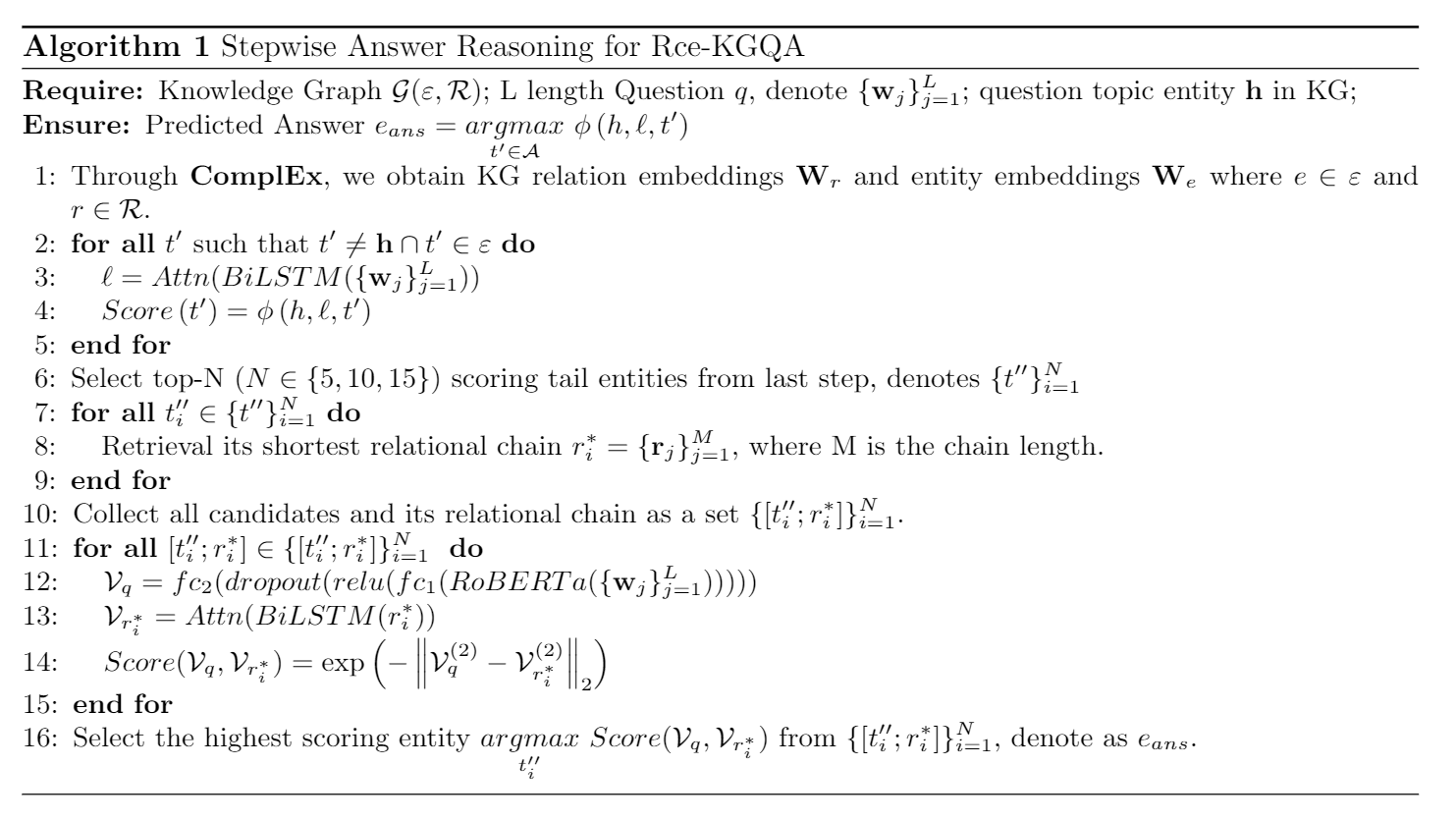}
\caption{The formulaic illustration of our Rce-KGQA model.}
\label{algorithm1}
\end{figure}

Before being fed into the next stage, we transform these intermediate candidate entities to their shortest relational chains which point to the question's topic entity by retrieving them in KG and mapping these ordered chains to sequences of embedding which correspond with our embedded KG. The \textit{Relational Chain Reasoning Module} receives the intermediate results generated by the last step. Then, it simultaneously utilizes the relational chain sequences and user-question to measure the mutual similarity score through our Siamese network. Taking the question relational chain reasoning details into consideration can help increase the accuracy of answer prediction compared with the first stage, the \textit{Answer Filtering Module}. Finally, after sorting the scored candidates, we choose the entity which has the highest similarity score as the final answer. \textbf{Fig.}\ref{algorithm1} formally illustrates the algorithm of how our method works and predicts the final answer for a given multi-hop question, where $\ell$ denotes the question semantic representation, and $\phi$ denotes the \textbf{ComplEx} scorer.

\subsection{Answer Filtering Module}
\label{entitydetails}

\begin{figure*}[t]
\centering
\includegraphics[width=1.0\linewidth]{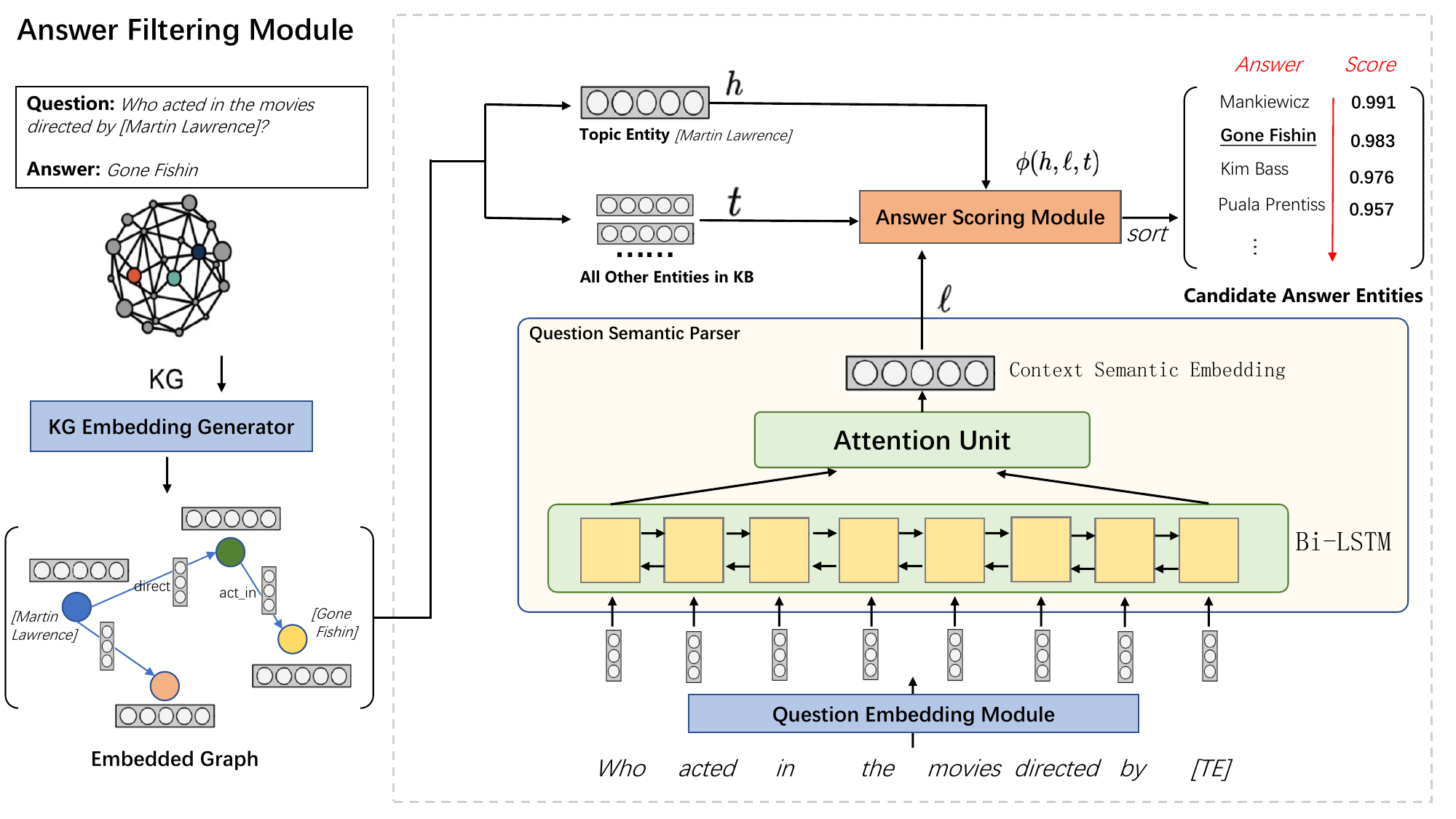}
\caption{Structure illustration of the Answer Filtering Module.}
\label{answer_filtering}
\end{figure*}

As the first step of our model, our \textit{Answer Filtering Module} aims to filter an entity set from all KG entities as candidate answers via three steps. These three operational steps are illustrated in Fig.\ref{answer_filtering} and relate respectively to three sub-modules: Graph Embedding Generator, Question Semantic Parser and Answer Scorer. We introduce each sub-module followed by the system processing order.

\subsubsection{Graph Embedding Generator}
\label{gem}
Traditional solutions could not handle many scenario problems such as \textit{Implicit Relation Reasoning} and \textit{Subgraph Neighborhood Constraint}. Inspired by the competitive performance of previous work EmbedKGQA \cite{apoorv2020embedKGQA}, we observe that the global relation knowledge and structure information preserved in KG embedding could potentially be used to resolve these issues efficiently and improve the overall accuracy of question answering.

In this work, our used KG is also embedded in continuous low-dimensional vector space to obtain all sparse representations for all entities and relations that existed in KG so we can simplify computations on the KG. We apply the \textit{Complex Embeddings} (ComplEx) \cite{trouillon2016ComplEx} approach to embed relations and entities in complex vector space. Compared with traditional KG embedding methods like \textit{TransE} \cite{bordes2013transe} and its extensions, semantic matching models such as Holographic Embeddings \cite{nickel2016holo}, ComplEx and RESCAL \cite{nickel2011rescal} have shown that they can generally yield better results. All KG embeddings are initialized randomly from uniform distributions. Generally, the hyper-parameters about entity and relation embedding dimension are not less than 100 and, in this paper, we set embedding dimension at 200 which follows previous similar works.

In each training step, positive facts which present correct real-world relational triples are sampled from all factoid triples existing in KG. Negative facts which present fake factoid relational triples are generated from negative sampling in the negative example generation step, where it randomly replaces the tail entity with an incorrect entity or replaces the relation with incorrect relation. 

Considering the samples count ratio of positive to negative ones, Trouillon et al. \cite{trouillon2016ComplEx} further investigated the influence of different numbers of negative samples for each positive one. Their work demonstrates that generating more negatives usually leads to better performance, and around fifty negatives per positive example is an appropriate trade-off between reasoning accuracy and training cost. So, our implementations also follow this prior setting.

Given $h, t \in \varepsilon$ and $\ell \in \mathcal{R}$, this embedding approach would provide $v_{h}, v_{\ell}, v_{t} \in \mathbf{C}^{d}$ for each relation triple $(h^{\prime}, \ell^{\prime}, t^{\prime})$, and the scoring function is defined as follows:

\begin{equation}
\label{equ_complex}
\begin{aligned}
\phi(h^{\prime}, \ell^{\prime}, t^{\prime}) &=\operatorname{Re}\left(\left\langle v_{h}, v_{r}, \bar{v}_{t}\right\rangle\right) \\
&=\operatorname{Re}\left(\sum_{k=1}^{d} v_{h}^{(k)} v_{r}^{(k)} \bar{v}_{t}^{(k)}\right)
\end{aligned}
\end{equation}

\begin{equation}
\label{filtering_equ1}
\phi\left(h^{\prime}, \ell^{\prime}, t^{\prime}\right)>0 \quad \forall a \in \mathcal{A}
\end{equation}

\begin{equation}
\label{filtering_equ2}
\phi\left(h^{\prime}, \bar{\ell^{\prime}}, \bar{t^{\prime}}\right)<0 \quad \forall \phi\left(h^{\prime}, \bar{\ell^{\prime}}, \bar{t^{\prime}}\right) \notin \mathcal{A}
\end{equation}

 where the $Re(\cdot)$ means taking the real part of a complex value , the $\bar{v}_{t}$ denotes the conjugate of ${v}_{t}$, the $\bar{\ell^{\prime}}, \bar{t^{\prime}}$ is the random replaced wrong relation and wrong tail entity of the $\ell^{\prime}, t^{\prime}$, and the $\mathcal{A}$ means the set including all real-world knowledge triples. 
 
 Optimization Eq.\ref{equ_complex} aims to minimize the values for all false triples less than 0,  Eq.\ref{filtering_equ2} and maximizes the values for all true triples greater than 0, Eq.\ref{filtering_equ1}. It can be easily carried out by stochastic gradient descent (SGD) or Adam optimizer at each training iteration. 
Lastly, the original structure and relation information in the KG are preserved in these learned vectors, which helps efficient completion of the downstream procedures. 

\subsubsection{Question Semantic Parser}
In this section, we introduce the \textit{Question Semantic Parser}, which consists of a recurrent neural network (bidirectional-LSTM) and extra self-attention operation, to help represent the question's meanings. During the inference procedure, the \textit{Question Semantic Parser} takes a question as the input and provides a predicted vector $\hat{\ell}$ as this question's relationship representation between the topic and answer in KG.
 
As shown in Fig.\ref{answer_filtering}, we build this sub-module based on the hierarchical neural network. Firstly, we encode the $L$ length question $\{t_{j}\}$, for $j = 1,...L$ into a sequence of word embedding vectors $\{v_{j}\}$ through our word embedding layer whose parameters are learnable during the training procedure. The word embedding dimension is consistent with the recurrent network's hidden dimension. Then we employ a single layer bidirectional LSTM to learn a forward hidden state sequence $\left(\overrightarrow{\mathrm{h}_{1}}, \overrightarrow{\mathrm{h}_{2}}, \ldots, \overrightarrow{\mathrm{h}_{L}}\right)$ and a backward hidden state sequence $\left(\overleftarrow{\mathrm{h}_{1}}, \overleftarrow{\mathrm{h}_{2}}, \ldots, \overleftarrow{\mathrm{h}_{L}}\right)$. Compared to general RNN, bidirectional LSTM is a special RNN which mainly solves the gradient disappearance and gradient explosion challenges and captures better long-distance semantics during long sequence modeling. Our LSTM component uses 256 as its dimension of hidden representations $h_{t}$ and memory cells $c_{t}$. It is well known that the performance of LSTMs depends crucially on their initialization and offers a strong starting point to facilitate model convergence, so we initialize our bidirectional LSTM weights with Xavier \cite{xavier} initialization which is markedly superior to other initialization methods such as Gaussian, Uniform and Kaiming initialization\cite{kaimingchushihua}.

Taking the forward step as an example, the next state ${\overrightarrow{\mathrm{h}_{j}}}$ based on last state ${\overrightarrow{\mathrm{h}_{j-1}}}$ is computed via the following operations.
\begin{equation}
\label{fgate}
    \mathbf{f}_{j}=\sigma\left(\mathbf{W}_{x f} \mathbf{x}_{j}+\mathbf{W}_{h f} \overrightarrow{\mathbf{h}_{j-1}}+\mathbf{b}_{f}\right)
\end{equation}

\begin{equation}
\label{igate}
    \mathbf{i}_{j}=\sigma\left(\mathbf{W}_{x i} \mathbf{x}_{j}+\mathbf{W}_{h i} \overrightarrow{\mathbf{h}_{j-1}}+\mathbf{b}_{i}\right)
\end{equation}

\begin{equation}
\label{ogate}
    \mathbf{o}_{j}=\sigma\left(\mathbf{W}_{x o} \mathbf{x}_{j}+\mathbf{W}_{h o} \overrightarrow{\mathbf{h}_{j-1}}+\mathbf{b}_{o}\right)
\end{equation}

\begin{equation}
\label{upcell1}
\mathbf{c}_{j}=\mathbf{f}_{j} \circ \mathbf{c}_{j-1}+\mathbf{i}_{j} \tanh \left(\mathbf{W}_{x c} \mathbf{x}_{j}+\mathbf{W}_{h c} \overrightarrow{\mathbf{h}}_{j-1}+\mathbf{b}_{c}\right)
\end{equation}

\begin{equation}
\label{upcell2}
\overrightarrow{\mathbf{h}_{j}}=\mathbf{o}_{j} \circ \tanh \left(\mathbf{c}_{j}\right)
\end{equation}

The variables $f_{j}, i_{j}, o_{j}$ in the above equations are the input, forget and output gate's activation vectors respectively, where $\mathbf{c}_{j-1}$ and $\mathbf{c}_{j}$ are the cell state vectors in the $j-1$ time and $j$ time, $tanh$ and $\sigma$ are the hyperbolic tangent and sigmoid functions.
Eq.\ref{fgate} denotes the forget gate operation, which aims to control whether to forget the hidden cell state's part information of the last moment with a certain probability. Eq.\ref{igate} denotes the input gate operation, which is responsible for processing the current sequence input. Eq.\ref{ogate} denotes the output gate operation, which determines the degree to which information is updated and output. Eqs. \ref{upcell1} and \ref{upcell2} are the steps to update the old cell state, which is determined jointly by the state of the previous sequence, this sequence's current input and the activation function.
After the information flows through LSTM, we concatenate the forward $\overrightarrow{\mathbf{h}_{j}}$ and backward $\overleftarrow{\mathbf{h}_{j}}$ and obtain the combined features $\mathbf{h}_{j} = [\overrightarrow{\mathbf{h}_{j}};\overleftarrow{\mathbf{h}_{j}}]$. After that, the last hidden state is considered to be the question semantic representation.
 
Different word tokens make different contributions to the relationship semantic recognition. For example, words which are prepositions and articles are more irrelevant for discovering question semantics than relational demonstrators. Thus, after LSTM we apply the self-attention mechanism to capture more valuable features. The attention operation details are shown in Eq.\ref{attenweight} and Eq.\ref{weightedvalue}. Given an LSTM hidden representation, a full connect layer, activation function $tanh$ and softmax operation will jointly generate the attention weight $\alpha_{j}$ at first. Then, as shown in Eq.\ref{weightedvalue}, the final attention vector representation $s_{j}$ which is the semantic representation of the language question is aggregated by the weighted sum operation of $\mathbf{h}_{j}$ and $a_{j}$.

\begin{equation}
\label{attenweight}
\alpha_{j}=\frac{\exp \left(a_{j}\right)}{\sum_{i=1}^{L} \exp \left(a_{i}\right)} \quad where\quad  a_{j}=\tanh \left(\mathbf{w}_{a}^{\top}\left[\mathbf{h}_{j}\right]+b\right)
\end{equation}

\begin{equation}
\label{weightedvalue}
s_{j}=\sum_{i} \alpha_{i j} \mathbf{h}_{i j}
\end{equation}

All the weight matrices, weight vector $\mathbf{W}$, and bias terms are calculated based on the training data,i.e. LSTM gate unit weight matrix $\{\mathbf{W}_{f},\mathbf{W}_{i},\mathbf{W}_{o}\}$ and attention weight matrix $\mathbf{w}_{a}^{\top}$. In this way, we obtain the rich relationship semantics implied in natural language questions for answer reasoning.

\subsubsection{Answer Scorer}
Like the ComplEx \cite{trouillon2016ComplEx} scoring function which depicted in Eqs. \ref{equ_complex}, \ref{filtering_equ1}, and \ref{filtering_equ2}, as shown in Eq.\ref{asm}, we learn an answer ranker $Rank\left(t\right)$ for each candidate $\textbf{t}$, namely \textit{Answer Scorer} to score the (topic entity, relationship semantic) pair against all possible KG entities $\textbf{t} \in \varepsilon$ by maximizing the probabilities of positive samples $t \in \mathcal{A}$ and minimizing the negative sample $t^{\prime} \notin \mathcal{A}$, where the $\mathcal{A}$ means the set including all real-world knowledge triples. 

\begin{equation}
\label{asm}
Rank\left(t\right)=\left\{\begin{array}{ll}
\operatorname{max}\left(\phi\left(h, \ell,t \right)\right), & \forall t \in \mathcal{A} \\
\operatorname{min}\left(\phi\left(h, \ell,t^{\prime} \right)\right), & \forall t^{\prime} \notin \mathcal{A}
\end{array}\right.
\end{equation}

Since our \textit{Question Semantic Parser} is designed to fit realistic relationship features, all the pretrained KG entity embeddings are frozen during the model convergence procedure.

Instead of simply selecting the entity with the highest score due to its low accuracy but high recall performance, we conduct a rough filtering by selecting top-n, where n $\in \{5, 10, 15\}$ to obtain the intermediate scored candidate entities that have a high answer recall rate. For the inference, the \textit{Answer Scoring Module} gives each candidate a plausibility score to indicate its answer confidence and filter out the top-n scored intermediate result which is fed into the next step, the \textit{Relational Chain Reasoning} module.

\subsection{Relational Chain Reasoning Module}
\label{relationdetails}

Our available KG often contains a large number of entities and has enormous factoid triples, and it could be inaccurate when comparing all candidate embedding representations against with each other. Specifically, after training the \textit{Answer Filtering Module}, we obtain all the scored entities for each training sample. During the prediction result analysis, we observe that the model performance on $hit@5$ outperforms that on $hit@1$ metric, which could be due to the influence of a large number of noisy entities number in the large-scale KG. Furthermore, since the answer is given as the only ground-truth information, a major challenge for multi-hop KGQA is that it usually lacks intermediate reasoning supervision signals.

\begin{figure*}[ht]
\centering
\includegraphics[width=1.0\linewidth]{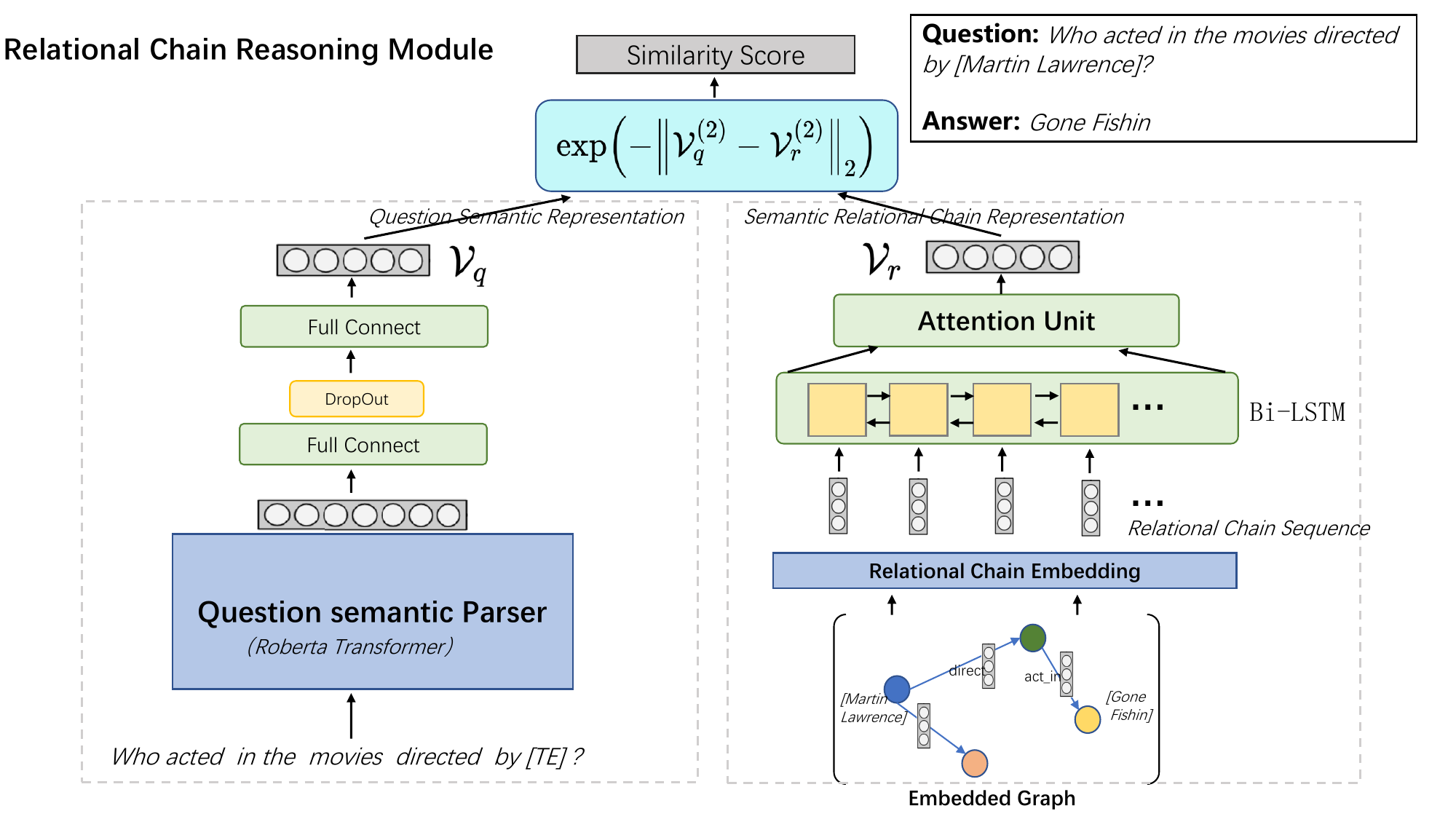}
\caption{Structure of the designed Relational Chain Reasoning Module.}
\label{relational_chain_reasoning}
\end{figure*}

To tackle these issues, we propose an extra component for our KGQA work, termed the \textit{Relational Chain Reasoning Module}. As the final and important step of our approach, it aims to improve the reasoning accuracy through considering the reasoning chain order and its relational type under a weak supervised situation. Its training procedure is irrelevant to the \textit{Answer Filtering Module}, but its training dataset is constructed from the \textit{Answer Filtering Module}'s prediction results.

Formally, we use the trained \textit{Answer Filtering Module} to obtain the sorted scored entities $\{[e_{it}; s_{it}]\}_{i=1}^{n}$, $\{[e_{iv}; s_{iv}]\}_{i=1}^{n}$ and $\{[e_{ie}; s_{ie}]\}_{i=1}^{n}$ for the training, validating, and testing datasets, respectively.
Then, these scored results are truncated by only reserving top-$n$ ($n \in \{5,10,15\}$) candidates. Next, looking at the rough filtered results in each experimental sample, if it belongs to the correct answers, we construct a corresponding positive sample $[\mathbf{q}; \{\mathbf{r}\}_{i=1}^{c}]$, in which $\mathbf{q}$ denotes the question tokens and $\{\mathbf{r}\}_{i=1}^{c}$ is the \textbf{c} length shortest path searched by the graph retrieval algorithm, as do the negative samples that are not inside in the correct answers.

\subsubsection{Siamese Network Based Similarity Scoring}
As shown in Fig.\ref{relational_chain_reasoning}, we introduce our novel sub-module \textit{Siamese Network Based Similarity Scoring}, which is essentially a Siamese network \cite{sia2016jonas}. The module's two feature detectors aim to extract the question Eq.\ref{rcrmq} and the relational chain semantic representations Eq.\ref{rcrmr}. They are constructed by a pretrained transformer \textbf{RoBERTa} \cite{roberta} to generate question vector $\mathcal{V}_{q}$ from question $q =\{\textbf{w}_{i}\}_{i=1}^{N}$ and a single-layer bidirectional \textbf{LSTM} to generate relational chain vector $\mathcal{V}_{r}$ from relational chain $r^{*} =\{\textbf{r}_{j}\}_{j=1}^{M}$. RoBERTa is a revolutionary self-supervised pretraining technique that learns to predict intentionally hidden sections of language text. Moreover, the representations learned by RoBERTa has been shown it can generalize outcomes superior even to many NLP downstream tasks compared to original BERT \cite{google2018bert}. Structure details will be described in the next successive two subsection. After information passes through these two encoded networks, the semantic feature similarities Eq\ref{rcrm} in the vector representation space are subsequently used to infer the semantic similarity between the question and relational chain from the topic entity to answer.

\begin{equation}
\label{rcrmq}
\mathcal{V}_{q} = W_{2}^{\top}(drop(W_{1}^{\top}(RoBERTa(q))+b_{1}))+b_{2}
\end{equation}

\begin{equation}
\label{rcrmr}
\mathcal{V}_{r} = Attn(BiLSTM(r^{*}))
\end{equation}

\begin{equation}
\label{rcrm}
 Score(\mathcal{V}_{q},\mathcal{V}_{r}) = \exp \left(-\left\|\mathcal{V}_{q}^{(2)}-\mathcal{V}_{r}^{(2)}\right\|_{2}\right)
\end{equation}

Here, we explain why we use a semantic similarity score between relation chain $\mathcal{V}_{r}$ and question representations $\mathcal{V}_{q}$ to determine the right answer in detail. Taking the above question ``Who acted in the movies directed by the director [Martin Lawrence]?'' as an example, we denote the natural language tokens as $q$. Inspired by literature in recent years such as \cite{pmlrv162das22a,fu2020surveyQA}, we intuitively 
suppose that the implied question semantic is similar and its representation is closer to the relational chain representation in vector space, ``\textit{directed\_by\_reverse} $\rightarrow$ \textit{starred\_actors\_reverse}'', which is correct in both order and type. It is inconsistent and far away from the relational chains which have the wrong type or order, such as ``\textit{starred\_actors\_reverse} 
$\rightarrow$ \textit{directed\_by\_reverse}'' and ``\textit{directed\_by\_reverse} $\rightarrow$ \textit{written\_by\_reverse}''.

As shown in Eq.\ref{rcrm}, we train the similarity scorer using the stochastic gradient descent (SGD) backward propagation algorithm under the mean-squared-error (MSE) loss function. Then we endow our training criterion with the $\ell^{2}$ (Euclidean distance) norm metric to avoid the model parameter distribution being highly warped. Regarding the predicted relatedness labels to lie in $[0, 1]$, for the positive sample we maximize the prediction value as close as possible to 1, and the negative sample as close as possible to 0. Finally, after sorting the scored candidates, we choose the entity which has the highest similarity score as the final answer.

\subsubsection{Question Semantic Representation}
As depicted in Fig.\ref{relational_chain_reasoning} left, we use and finetune a standard version of \textbf{RoBERTa} to obtain the hidden state $\mathcal{V}_{q}$ lying in start token $[CLS]$ for our question encoders. Note that we reformat question $q$ through replacing the topic entity's mention in the question with a token ``NE''. And we respectively supplement two special characters `$[CLS]$' and `$[SEP]$' before and after the reformatted question, which follows the BERT default configuration. This aforesaid operations can help our model better distinguish the topic entity and other question words mentioned. Afterwards, we link the question's topic entity mention to the KG node through matching with standard KG entity literal representations. We regard it as the semantic signal for answer reasoning and then adopt two full connect layers, neuron activation function ReLU \cite{relupaper} and a dropout layer to get better feature learning capability. After the above step, a vector representation is generated by the last full-connected layer, and we use the vector to compare space distance similarity with the output of another feature encoder, \textit{Relational Chain Representation} Module.

\subsubsection{Relational Chain Representation}
As depicted in Fig.\ref{relational_chain_reasoning} right, we provide the Relational Chain Representation module which consists of a single-layer bidirectional LSTM and a self-attention layer. Given the KG relational chain embedding sequence as input, this module learns and captures the relevant and necessary semantic information for answering reasoning. It has a similar structure to the \textit{Question Semantic Parser} in the \textit{Answer Filtering Module}, but the token embeddings used are not initialized randomly in this case. Instead, we apply the pretrained KG relation embedded representations existing in Sec ~\ref{gem} to embed our relational chain.

Formally, as depicted in Eq.\ref{rcrmr}, for relational chain $\mathbf{r}^{*} =\{\textbf{r}_{j}\}_{j=1}^{M}$, where $M$ denotes the chain length, each relationship representation $\textbf{r}_{j}(j = 1,2,...,M)$ is initialized with the pretrained KG relational embeddings. Next, we feed the embedded vectors $\mathbf{r}^{*}$ into a single-layer bidirectional LSTM network to obtain a series of output states $h_{1}, h_{2},... , h_{M}$, where the \textit{j}-th relation $h_{j}$ denotes $[\overrightarrow{\mathbf{h}_{j}};\overleftarrow{\mathbf{h}_{j}}]$, a combined vector of forward LSTM state output $\overrightarrow{\mathbf{h}_{j}}$ and backward LSTM state output $\overleftarrow{\mathbf{h}_{j}}$. Then the new question representation $q = [\mathbf{h}_{1}, \mathbf{h}_{2},... , \mathbf{h}_{M}]$ can be transformed through a self-attention operation, which is similar to \textit{Answer filtering Module} counterpart, shown in Eq.\ref{attenweight}, \ref{weightedvalue}. Through this forward propagation, that is similar to \textit{Question Semantic Parser}, we can obtain a question semantic vector $\mathcal{V}_{q}$ which has the same dimension as relational chain vector $\mathcal{V}_{r}$ for the following similarity computation step.

\section{Experiment}

\label{experiment}
In this section, we evaluate our proposed Rce-KGQA against competitive baselines on three benchmark datasets to investigate whether our model can outperform other methods on reasoning over the weakly supervised signal and incomplete knowledge graph. And we also append extensive ablation experiments and a case study to carefully verify and vividly demonstrate that the necessity, superiority, and meaning about our ideas in this work. The datasets as well as the pytorch implementation of our model are publicly available at \url{https://github.com/albert-jin/Rce-KGQA}.

\subsection{Datasets and Evaluation Metric}

Here, we first describe the three benchmark datasets we use in this work and then give a brief introduction of the metric $hit@1$ we used for model evaluation.

\textbf{MetaQA} \cite{zhang2018metaQA} is a large KGQA dataset which provides an original version \textit{Vanilla} and two variations. In this paper, we use the original ones because they are designed manually. This dataset includes up to 75w QA pairs which are merely 2-hop questions. The questions' literal descriptions are generated by cross-language translation, \textit{English$\rightarrow$French$\rightarrow$English}. This relies on a large-scale movie domain which contains 9 relationship types, 43234 entities and up to 135k factoid triples. Here, we use the dataset edition which is generated from Apoorv et al. \cite{apoorv2020embedKGQA}.

\textbf{WebQuestionsSP-tiny} \cite{tauyih2016wbquestion} dataset is a relatively small dataset with a total of 4736 QA pairs. This QA dataset's available KG is a subset of Freebase  that contains all the facts within 2-hops of any entity mentioned in the questions of the original WebQuestionsSP, which have more than 188w entities and 1000 relationship types. Following \cite{apoorv2020embedKGQA}, for all topic entities labeled in the original Freebase, He et al. \cite{he2021stu-teaQA} construct a subgraph containing other KG entities close to them by the PageRank-Nibble algorithm (PRN) \cite{rpn}. In this way, theoretically, the pruned KG is likely to contain the corresponding answer entity for close to whole questions. In this work, we use the same train/dev/test splits as GraftNet \cite{sun2018graftnet}.


\textbf{Complex WebQuestionsSP} (Complex-WebQSP)\cite{TalmorCWQ} is a more complex multi-hop reasoning dataset and its questions require up to a 4-hops relational path during answer reasoning. There are four types of question: conjunction(about 45\%),  composition(about 45\%), comparative (about 5\%), and superlative (about 5\%). Our used edition of the Complex-WebQSP dataset is obtained from He et al. \cite{he2021stu-teaQA}.

\begin{table}[h]
\renewcommand{\arraystretch}{1.2}
\caption{Statistics for dataset MetaQA, WebQuestionsSP-tiny and Complex WebQuestionsSP. MetaQA contains three subsets of different complex question relational chain length, \textbf{1}/\textbf{2}/\textbf{3} hop MetaQA. WebQuestionsSP-tiny dataset requires up to 2-hop reasoning from knowledge base. Meanwhile, the Complex-WebQSP dataset requires up to 4-hops of reasoning on the KG.}
\label{dataset_details}
\centering
\begin{tabular}{c|ccc}  \hline
Datasets & Train & Dev & Test \\ \hline
MetaQA 1-hop & 208970 & 19947 & 9992 \\
MetaQA 2-hop & 231844 & 14872 & 14872 \\ 
MetaQA 3-hop & 227060 & 14274 & 14274 \\ \hline
WebQSP-tiny & 2848 & 250 & 1639 \\ 
Complex-WebQSP & 27639 & 3519 & 3531 \\ \hline
\end{tabular}
\end{table}

\textbf{Metric} $hit@1$ is a standard assessment for measuring the ratio in all validation samples that the highest scored entity belongs to the correct answers. In brief, if the QA system provides the user with a single entity and this entity is right, we then determine that this prediction is correct. This evaluating indicator is popular and publicly recognized and has been used in many recent KGQA works \cite{apoorv2020embedKGQA,chen2019uhopKGQA,yunshi2020graphKGQA}.

\subsection{Experiment Setting}
As we know, hyperparameter choices have a significant impact on the model's final performance \cite{electronics3zuojwq,cvtassd1zuojwq,zzm2022ijannconfere}. Our optimal model hyperparameter configuration is summarized as follows. All the LSTM modules we used as feature encoders have a single layer with a hidden dimension of 256. All the dropout layers randomly drop about 30\% of their features for inputs during the training step but they do not drop any features during testing. We apply Xavier initialization to each network layer's training parameters in our model.
Our applied pretrained transformer \textit{RoBERTa} is a PyTorch-implemented configuration, which uses the BERT-base architecture \cite{tqx1zuoenhancere,ghj2022KSEMconfere,google2018bert}, consisting of 12 layers,768-d hidden size and 12 attention heads for efficient training and inference. Roberta-base encoder \cite{ott2019fairseq,roberta} contains up to about 125M parameters. The Answer Filtering Module is trained for up to 200 epochs with a batch size of 128, and the relational chain reasoning module is trained for up to 120 epochs with a batch size of 32 on three benchmarks. For every 10 training epochs, we adopt the early-stopping strategy by evaluating $hit@1$ on the test set to avoid overfitting. During model convergence, the stochastic gradient descent (SGD) optimizer with initial learning rate \textit{lr} = 1e-5 is adopted. We used different random seeds to validate our best-configured model independently 5 times and report the average validation performances of our model in the next sections.

\subsection{Compared Methods}
In our experiment, the state-of-the-art methods for comparison are described as follows:

\begin{itemize}
\item \textbf{EmbedKGQA} \cite{apoorv2020embedKGQA} is a KG embedding driving method for multi-hop KGQA which matches the pretrained entity embeddings with question embeddings generated from the transformer.
\item \textbf{SRN} \cite{srn} is an RL-based multi-hop question answering model which conducts the QA task by extending the inference chains on a KG.
\item \textbf{KVMem} \cite{kvmem} The Key-Value Memory Network first attempts to conduct QA over incomplete KGs by augmenting it with text. It uses a memory table which stores the KG facts encoded into key-value pairs to retrieve a question-specific subgraph for reasoning. 
\item \textbf{GraftNet} \cite{sun2018graftnet} is a question description-based semantic sub-graph driving method that uses a variational graph CNN to perform QA tasks over question-specific subgraphs containing KG facts, entities and discourses from textual corpora.
\item \textbf{PullNet} \cite{pullnet} improves GraftNet on the retrieval subgraph by introducing the graph retrieval module which utilizes shortest path from the topic entity to answer as the additional supervised signal.

\item \textbf{NSM}$_{s}$ \cite{he2021stu-teaQA} is a series of teacher-student learning approaches implemented as based on the Neural State Machine \cite{nsm}. $NSM$, $NSM_{+p}$, and $NSM_{+h}$ are three model variants which effectively employ the intermediate supervision signals. Specifically, $NSM$ do not use the teacher network, $NSM_{+p}$ use the teacher network with parallel reasoning, and $NSM_{+h}$ use the teacher network with hybrid reasoning.
\end{itemize}

\subsection{Main Results}
In this section, we compare our model with the state-of-the-art baseline methods on three benchmarks, and the following questions are answered:

\textbf{Q1.}\quad How effective and accurate is the performance of our model compared with other SOTA models?

\textbf{Q2.}\quad Can our model really identify the implicit relations and the indirectly linked answer when there is no direct relational chain from topic entity to answer?

\textbf{Q3.}\quad How many parameters does our \textit{Rce-KGQA} contain, and does our model have a certain high execution efficiency?

\subsubsection{Answer Reasoning on MetaQA}
\label{ar_metaqa}
As illustrated in Table~\ref{comparison_sota_metaqa}, the overall experimental results on the \textit{MetaQA} test set clearly demonstrate that our proposed KGQA architecture significantly outperforms state-of-the-art methods on $hit@1$ metric. Specifically, according to the performance of our model's result on 1-hop MetaQA (identical to WikiMovies), compared to the other methods, Row \textbf{1} \textasciitilde \textbf{5}, we observe that our method's $hit@1$ accuracy achieves much higher performance up to \textbf{98.3}\%, which is an increase of \textbf{1.3}\% compared to \textbf{GraftNet} \cite{sun2018graftnet} and \textbf{PullNet} \cite{pullnet}, an increase of \textbf{0.8}\% compared to \textbf{EmbedKGQA} \cite{apoorv2020embedKGQA} and an increase of \textbf{1.1}\% compared to \textbf{NSM}$_{+h}$ \cite{he2021stu-teaQA}.

For the evaluation of multi-relation questions which require at least two hops of inference to find the answers, $hit@1$ results on 2-hop and 3-hop MetaQA also show better performance than most competitive state-of-the-art baselines. Although our model test results on 2-hop did not achieve the best score, it is nevertheless comparable to other SOTA models. It is worth noting that the retrieval-and-reason process of \textbf{PullNet}, which can simultaneously extract answers from both corpora and KGs, is good at reasoning answers over large-scale KGs such as MetaQA. In contrast, we can see that our model \textit{Rce-KGQA}, which takes relation chain reasoning into consideration, does not drop significantly in performance but remains almost unchanged when the relational chain hop increases. We think the reason may be that the baseline models only consider the question shadow semantic representations, and inevitably introduce noise and incorrect retrieving path over KG. On the other hand, since our model focuses on relational chain order and relation type, it is less sensitive to the hop size and shows robustness on complex multi-constraint queries over KG. 

In summary, these results have shown their effectiveness and superiority when considering the question semantic and its relational chain into reasoning, which largely improves the KGQA performance. 

\begin{table}[h]
    \centering
    \begin{tabular}{c|ccc} \hline
        Model & 1-hop MetaQA & 2-hop MetaQA & 3-hop MetaQA\\ \hline
        EmbedKGQA & \underline{97.5} & 98.8 & 94.8 \\
        SRN & 97.0 & 95.1 & 75.2 \\
        KVMem & 96.2 & 82.7 & 48.9 \\
        GraftNet & 97.0 & 94.8 & 77.7 \\
        PullNet & 97.0 & \bf{99.9} & 91.4 \\ \hline
        NSM & 97.1 & \bf{99.9} & \bf{98.9} \\ 
        NSM+p & 97.3 & \bf{99.9} & \bf{98.9} \\ 
        NSM+h & 97.2 & \bf{99.9} & \bf{98.9} \\ \hline
        Our Model  & \bf{98.3} & \underline{99.7} & \underline{97.9} \\ \hline
    \end{tabular}
    \caption{Effectiveness comparisons on three subsets of MetaQA. The first group of results was taken from papers on recent methods. The values are reported using hits@1. The number in \textbf{bold} and \underline{underlined} number denote the best and second-best methods, respectively. This figure corresponds to Sec. \ref{ar_metaqa}.}
    \label{comparison_sota_metaqa}
\end{table}


However, we also consistently find that our proposed Rce-KGQA's two-stage pipeline mechanism could bring deviation cascade propagation between the coarse-grained answer filtering procedure [\ref{answer_filtering}] and the fine-grain answer selecting procedure [\ref{relational_chain_reasoning}]. As is shown in Table~\ref{comparison_sota_metaqa}, and the comparison results from the columns of 2/3-hop MetaQA indicated. Our approach performs poorly when compared with other KGQA models such as \textbf{PullNet} and \textbf{NSMs}. They all consistently achieved very high performance with the $hit@1$ metric of 99.9 percentage on 2-hop MetaQA and 98.9 percentage $hit@1$ on the 3-hop MetaQA dataset. Correspondingly, our Rce-KGQA underperforms with about 0.2 percentage points behind on the 2-hop MetaQA and about 1.0 percentage points behind on the 3-hop MetaQA dataset. 


In general, although our two-stage pipeline solution outperforms many baselines such as GraftNet and PullNet \cite{sun2018graftnet,pullnet}, our designed Rce-KGQA's separate architecture determines the fact that the quality of the final answer provided by \textit{Relational Chain Reasoning Module} completely depends on the quality of the candidate entities provided by \textit{Answer Filtering Module}. The feature of pipeline architecture like this is the principal reason that inevitably causes the cascaded error propagation, which would bring down the overall performance of the question answering service. We think it is crucial to enhance our model \textit{Rce-KGQA} by integrating these two separated modules [\ref{answer_filtering}, \ref{relational_chain_reasoning}] into one joint module, which we leave for future work. 

In addition to the embedded graph vector's parameter volumes, our two-stage modules respectively contain 46M parameters (mainly owned by \textit{BiLSTM-Attn} block) and 197M parameters (mainly owned by \textit{Roberta} encoder and \textit{BiLSTM-Attn} block). Through our experiments, we observe that the total inference time fluctuates between 1.9 seconds and 2.7 seconds, in which the \textit{Answer Filtering Module} contributes about 0.6s and the \textit{Relational Chain Reasoning Module} contributes about 1.4s. It is conceivable that if the encoder were switched to RoBERTa using the BERT-large architecture \cite{ott2019fairseq,roberta}, it would mean more heavy parameters, including up to 355M parameters and additional training optimization procedures and more inference time-consuming. Based on the experimental reproducibility and the consideration of the model lightweight, we chose the basic configuration of the \textit{Roberta} encoder.

\subsubsection{Experiments on WebQSP-tiny and Complex-WebQSP} 
\textit{WebQuestionsSP-tiny} \cite{pmlrv162das22a,he2021stu-teaQA,apoorv2020embedKGQA} is a relatively small dataset for training but relies on a large-scale KG (Freebase) whose entities' count is greater than 10 million. Table~\ref{comparison_sota_wbq} presents the evaluation results on the \textit{WebQuestionsSP-tiny} validation dataset, from which we can observe that our KGQA system still performs better that other state-of-the-art counterparts, \textbf{EmbedKGQA} (has 3.8\% lower hists-at-one than our model) and \textbf{PullNet} (has 2.3\% lower hists-at-one than our model). 

\begin{table}[h]
    \centering
    \begin{tabular}{c|c|c} \hline
    Model & WebQuestionsSP-tiny & Complex-WebQSP\\ \hline 
    KVMem & 46.7 & 21.1 \\ \hline
    GraftNet & 66.4 & 32.8 \\ \hline
    EmbedKGQA & 66.6 & - \\ \hline
    PullNet & 68.1 & 45.9 \\ \hline
    NSM & 68.7 & 47.6 \\ \hline
    NSM+p & \underline{73.9} & \underline{48.3} \\ \hline
    NSM+h & \bf{74.3} & \bf{48.8} \\ \hline
    Our Model & 70.4 & \underline{48.3} \\ \hline
    \end{tabular}
    \caption{Experiment results (\% Hits@1) compared with SOTA methods on the WebQuestionsSP-tiny and Complex WebQuestionsSP validation datasets. All QA pairs in WebQuestionsSP-tiny are 2-hop relational questions. We copy the results for KV-Mem, GraftNet, EmbedKGQA, PullNet and NSM from \cite{kvmem,sun2018graftnet,apoorv2020embedKGQA,pullnet,he2021stu-teaQA}, respectively. The best score is in \textbf{bold} and the second-best score is \underline{underlined}.}
    \label{comparison_sota_wbq}
\end{table}

Specifically, the last row shows that our full model achieves accuracy of up to 70.4\% $hit@1$, which improves a large margin to other prior models. A possible explanation is that the filtering model equips the extra relational chain module with better reasoning perception, leveraging KG and question implicit features more efficiently, and emphasizing the order of relational triples selection to help our model make a correct decision. Even in large-scale KGs along with small training dataset situations like \textit{WebQuestionsSP-tiny}, our \textit{Rce-KGQA} solution can still be robust and helpful for handling realistic QA applications.

\textit{Complex-WebQSP} \cite{he2021stu-teaQA} is a derivative KGQA dataset edition which is generated from \textit{WebQuestionsSP-tiny} by extending the question entities or adding constraints to answers. As its name indicates, most of the questions this dataset included require up to 4-hops of relational chain reasoning from the topic entity to the corresponding answers.
 
The third column of Table~\ref{comparison_sota_wbq} reports the $hit@1$ metric statistics on the Complex-WebQSP benchmark. Our model outperforms competitively with state-of-the-art KGQA baselines on such a complex multi-hop question answering scenario. More specifically, among most baselines (KVMem $\thicksim$ NSM), our Rce-KGQA significantly surpasses other baselines, and respectively achieves 27.2\%, 15.5\%, 3.3\%, and 0.7\% absolute gains over these baselines (KVMem $\thicksim$ NSM) in terms of the overall metric $hit@1$. This establishes the fact that our \textit{Rce-KGQA} is better than previous approaches in terms of answering the questions with long-distance relational dependency. The \textbf{NSM}$_{+h}$ achieves the best performance on the two adopted benchmarks. The \textbf{NSM}$_{+p}$ and our \textit{Rce-KGQA} both gain the second best performance on the \textit{Complex-WebQSP} benchmark, proving both our model's competitive capability and the importance of the added teacher network. This is an important observation and advancement when it comes to handling such complex question answering tasks since our proposed approach is robust and efficient in dealing with these complex questions in multi-relational-hop answer reasoning situations.


\subsection{Answer Reasoning for implicit relationship discovery}

As shown in Table~\ref{comparison_full_half}, we verify our method's ability to discover missing implicit relationships through comparison experiments. The KG which MetaQA uses has no missing link during the reasoning path because the QA question pairs are constructed upon this KG. However, to make it become a realistic setting, we simulate an incomplete KG by randomly removing half (with probability = 0.5) of the factoid triples from it.  We call this pruned setting \textbf{half} and we call the full KG setting \textbf{full} in the text.  

The experiments show our method's implicit relation discovering capability substantially outperforms other state-of-the-art methods  over incomplete KGs. The amount of improvement is significant, with  an increase of 43.7\% compared to \textbf{KVMem} in $hit@1$. Furthermore, our competitive model also delivers an average 1.7\% $hit@1$ rate on 2-hop MetaQA half setting and performs well on 3-hop MetaQA half setting while \textbf{PullNet} still achieves the highest $hit@1$ score.

Hence many baseline methods such as \textbf{GraftNet}, \textbf{PullNet} require constructed question-specific subgraphs, indicating they lack the capability to recall the answer nodes out of their generated subgraph and cannot perform well in real QA scenarios. Fortunately our model, which exploits the KG link prediction properties, does not limit its capability due to this constraint. Although those complex questions in \textit{WebQuestionsSP-tiny} could be easily covered by hand-crafted rules, as many have been, our model is not suitable for such pre-defined rules. We think it is crucial to use more advanced reasoning capabilities to enhance our model \textit{Rce-KGQA} correctly, a task which we leave for future work.

\begin{table}[h]
    \centering
    \begin{tabular}{c|c|c|c|c} \hline
        \multirow{2}{*}{Models} & \multicolumn{2}{c}{2-hop MetaQA} & \multicolumn{2}{c}{3-hop MetaQA}\\ \cline{2-5}
         & \textbf{full} & \textbf{half} & \textbf{full} & \textbf{half} \\ \hline
        KVMem & 82.7 & 48.4 & 48.9 & 37.6 \\
        GraftNet & 94.8 & 69.5 & 77.7 & 66.4 \\
        PullNet & \bf{99.9} & 90.4 & 91.4 & \bf{85.2} \\
        EmbedKGQA & 98.8 & \underline{91.8} & \underline{94.8} & 70.3 \\ \hline
        Our Model  & \underline{99.7} & \bf{92.1} & \bf{97.9} & \underline{84.7} \\ \hline
    \end{tabular}
    \caption{Experimental results about reasoning on incomplete KG ($hit@1$ as a percentage). We consider two different KG settings, \textbf{full} and \textbf{half}. \textbf{Full} denotes the complete KG and \textbf{half} denotes a KG subset whose 50\% factoid triples are randomly removed.}
    \label{comparison_full_half}
\end{table}

\subsection{Answer Filtering Result Analysis}
To further examine whether our proposed enhancement to the extra module \textit{Relational Chain Reasoning Module} with advanced and obvious improvements, we analyze the reasoning performance of first module \textit{Answer Filtering Module} and show the answer distributions with prediction in Table~\ref{comparison_hit1}. In Table~\ref{comparison_hit1}, as we can see that, if we regard the scored candidate entities provided by this module as the final answer, the $hit@1$ accuracy rate drops markedly compared to $hit@5$ and $hit@10$. This observation indicates that the model which does not consider relational chain order and relation type achieves very poor performance and proves the necessity and superiority of our proposed module, the \textit{Relational Chain Reasoning Module}.

Furthermore, from Table~\ref{comparison_hit1}, we can clearly observe that almost right answers of our used datasets are collectively distributed in the top-5 of our model's predictions. Due to the high recall rate of our first module, the \textit{Answer Filtering Module}, we think we can only rely on a few top-scoring candidates (such as top-15, top-10, or even top-5) to further filter the final answer more accurately. So, during our model training and inference experiments, we tried several experimental schemes and considered the number of top-scoring candidates specifically, how to select candidates to automatically generate positive or negative samples and how to cut the top-N candidates for the sub-module \textit{Relational Chain Reasoning Module} inference. The related experimental details are shown in Sec.~\ref{topNanalysis}.

\label{topNanalysis}

\begin{table}[h]
    \centering
    \begin{tabular}{c|c|c|c} \hline
        Dataset & $Hit@1_{-r}$ & $Hit@5_{-r}$ & $Hit@10_{-r}$ \\ \hline
        2-hop MetaQA  & 0.861 & 0.995 & 0.999 \\
        3-hop MetaQA  & 0.858 & 0.984 & 0.997 \\ \hline
    \end{tabular}
    \caption{Our \textit{Answer Filtering Module} answer reasoning performance on three different metrics $Hit@1, 5, 10$. Model performance on $Hit@5$ and $Hit@10$ accuracy highly outperform over $Hit@1$ accuracy.}
    \label{comparison_hit1}
\end{table}

\subsection{Candidates Filtering Strategy Analysis}
Our QA solution \textit{Rce-KGQA} is a complex pipeline system, and we inevitably must choose some crucial hyper-parameters to acquire an optimal model. These pre-defined parameters include full connected/LSTM layer number, dropout rate, learning rate, and so on. As illustrated in Sec.~\ref{topNanalysis} and Table~\ref{comparison_hit1}, different selection modes about top-scoring candidates during training and inference have a huge impact on the final model performances. We now further investigate what influence would our model experience in different candidates selection mode.

Firstly, from Table~\ref{comparison_hit1} we can observe that our first step `answering filtering' consistently achieves high recall performances on $hit@5$. Now we manually choose and cut-difference sorted candidate answers and conduct our comparison experiments. Specifically, from Sec.~\ref{relationdetails} we know that the dataset for \textit{Relational Chain Reasoning Module} training is dynamically constructed by the last step. And the data construction follows the selection of top-\textbf{N} sorted scoring entities in which number \textbf{N} simultaneously decides the proportion of positive/negative samples and the intermediate results selection during the inference step. For example, in question \textbf{Q}, during model evaluation, we firstly obtain the intermediate scoring candidate answers $\{[A_{i};S_{i}]\}_{i=1}^{N}$. In the next step, we should use the \textit{Relational Chain Reasoning Module} to provide all filtered candidates with a more precise score to reach a final answer. Selecting how many top scoring entities to conduct the further step precisely becomes our focus research point.

We choose four strategies which include top-\{5, 10, 15, 20\}; then, from the corresponding experiments, we receive the following results which is shown in Table~\ref{topselectionpolicy}. From experimental comparison results we clearly find that in top-5 selection strategy, our model achieves the highest $hit@1$ performances and achieves the second highest performances in the top-10 strategy, in both 2-hop MetaQA and 3-hop MetaQA datasets. These phenomena can come from two aspects. First, the recall rate of the module \textit{Relational Chain Reasoning Module} is good enough for the next fine-grained screening and the positive/negative samples proportion should be in a suitable extent. In addition, more candidates in the model inference step could introduce more noise entities which could affect model answer judgment and decrease the overall model performances.

\begin{table}[h]
    \centering
    \begin{tabular}{c|c|c|c|c} \hline
        Policy & top-5 pick & top-10 pick & top-15 pick & top-20 pick \\ \hline
        2-hop MetaQA  & \textbf{0.997} & \underline{0.994} & 0.989 & 0.985 \\
        3-hop MetaQA  & \textbf{0.979} & \underline{0.971} & 0.967 & 0.967 \\ \hline
    \end{tabular}
    \caption{Our model final performance statistics about impacts of the four selection strategies: choose top-5, choose top-10, choose top-15 and choose top-20. The values reported are $hit@1$. Bold and underlined fonts denote the best and the second-best selection strategies.}
    \label{topselectionpolicy}
\end{table}

\subsection{Ablation Study}
To better understand and gain a deep insight into our model design, we also perform ablation experiments to investigate systematically the impact and contributions of different components.  \textbf{RceKGQA$_{-r}$, RceKGQA$_{-a}$} and \textbf{ RceKGQA$_{-b}$} are variants of our full model, RceKGQA. Note that, for our ablated experiments, we remove one component each time. Here we briefly introduce these variants for the ablated experiments.
\begin{itemize}
\item \textbf{RceKGQA$_{-r}$} removes the \textit{Relational Chain Reasoning Module} and the highest scoring entity is provided by the \textit{Answer Filtering Module} as the final answer. 
\item \textbf{RceKGQA$_{-a}$} removes all the self-attention operations from the model. 
\item \textbf{RceKGQA$_{-b}$} replaces RoBERTa with LSTM in the question semantic representation part of the \textit{Relational Chain Reasoning Module}.
\item \textbf{RceKGQA} is the full model introduced in this paper.
\end{itemize}

In this section, the following questions are answered:

\textbf{Q1.}\quad How much does the \textit{Relational Chain Reasoning Module} help our model's reasoning accuracy?

\textbf{Q2.}\quad Can the attention mechanism really help increase our model's overall performance?

\textbf{Q3.}\quad Is the effectiveness of our method due to the use of \textbf{RoBERTa} in the \textit{Relational Chain Reasoning Module}?

According to the result comparison between RceKGQA and its variant RceKGQA${-r}$
, we can clearly conclude that removing the \textit{Relational Chain Reasoning Module} from the proposed model has a huge impact on the results. 

The performance gap between RceKGQA${-r}$ which is shown in Row \textbf{2} and our full model as shown in Row \textbf{1} indicates that the semantic relational chain factor plays a pivotal role in answer reasoning, which incorporates relational chain order and relationship type to provide more accurate answers for the terminal user.

As shown in Row \textbf{3}, when our full model is compared with RceKGQA$_{-a}$, we can see an average 3.5\% performance drop across the $hit@1$ metric if the self-attention components are removed, demonstrating the importance of the self-attention mechanism used in our method, as it effectively helps with the final answer prediction. A possible reason is that the self-attention mechanism can distinguish the most relevant and interesting signals from noise information, and help our model better understand the question and relational chain semantic.

RceKGQA${-b}$, which is shown in Row \textbf{4}, only loses 1.6\% $hit@1$ accuracy compared with our full model, which replaces RoBERTa with BiLSTM, and demonstrates that using the transformer is not a major factor in increasing the overall model performance. 

The above-ablated statistics confirm that all three components introduced for handling the multi-hop relation question answering contribute to the overall model performance.

\begin{table}[h]
    \centering
    \begin{tabular}{c|ccc} \hline
        Model & 1-hop MetaQA & 2-hop MetaQA & 3-hop MetaQA \\ \hline
        RceKGQA & \bf{98.3} & \bf{99.7} & \bf{97.9} \\ \hline
        RceKGQA$_{-r}$ & 85.8 & 86.1 & 84.8 \\
        RceKGQA$_{-a}$ & \underline{96.1} & 95.9 & 93.4 \\ 
        RceKGQA$_{-b}$ & 95.9 & \underline{98.2} & \underline{95.6} \\ \hline
    \end{tabular}
    \caption{Ablation study statistical results for RceKGQA and its three variants ($Hit@1$ by percentage). Compared with our full model, suffix $_{-r}$ denotes RceKGQA whose \textit{Relational Chain Reasoning Module} is removed, suffix $_{-a}$ denotes the RceKGQA variant whose attention operation is dropped and suffix $_{-b}$ denotes the RceKGQA variant whose question encoder RoBERTa is replaced with BiLSTM.}
    \label{comparison_ablation}
\end{table}

\subsection{Case Study}
The major novelty of our approach lies in the introduced \textit{relational chain reasoning} network. Here, we present a case study to demonstrate its contribution to improving the overall model architecture.


 \begin{figure}[h]
\centering
\includegraphics[width=1.0\linewidth]{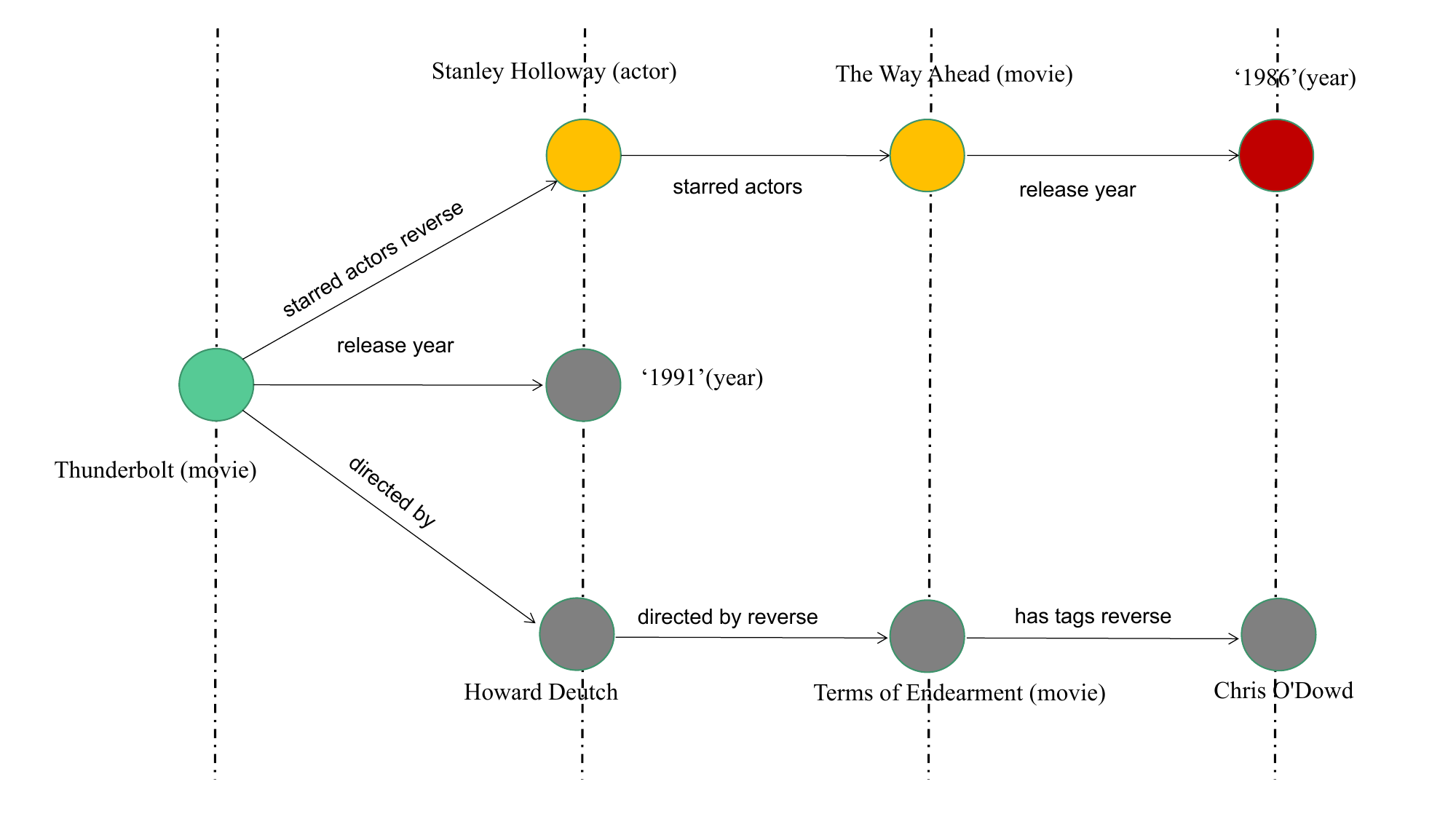}
\caption{ Case analysis from the 3-hop MetaQA dataset. We use green, red, yellow and grey circles to denote the topic KG nodes, correct answer, intermediate nodes and irrelevant nodes, respectively. The orange and red coloured circles denote the actual reasoning intermediate nodes and answer nodes. The colour darkness indicates the relevant degree of an entity by a method.}
\label{case_study}
\end{figure}

As shown in Fig. \ref{case_study}, 
given the question, ``\textit{In which years were movies released which starred actors who appeared in the movie [Thunderbolt]?}'', the right reasoning relational chain preserved in KG is \textbf{Thunderbolt}(movie)-\underline{starred\_actors\_reverse} $\rightarrow$ \textbf{Stanley Holloway}(actor)-\underline{starred\_actors} $\rightarrow$ \textbf{The Way Ahead}(movie)-\underline{release\_year} $\rightarrow$ `\textbf{1986}'(year). When ignoring the relational chain feature factor and only using the high-scored entity generated by the \textit{Answer Filtering Module} as the final answer, the network mistakenly selects a wrong reasoning path \textbf{Thunderbolt}(movie)$-$\underline{release\_year} $\rightarrow$ `\textbf{1991}'(year) for the aforesaid question with a very high probability of \textbf{0.96} as the answer. Its attention only focuses on the relationship: \underline{release\_year}, which ignores the repeated relations of \underline{starred\_actors} and \underline{starred\_actors\_reverse}. In comparison, the complete model which considers the relational chain factor and utilizes the \textit{relational chain reasoning} network for fine-grained selection can easily and correctly provide the right answer `\textbf{1986}'(year) with a high probability of \textbf{0.99} from KG.

This example shows that our \textit{relational chain reasoning} network indeed provides very useful supervision signals of relational chain recognition at intermediate steps to improve our model's overall QA performance.
\label{casestudy}

\section{Conclusion and Future Perspective}
\label{conclusion}
In this work, we introduce an elaborate KG embedding-based pipeline approach for the multi-hop KGQA task, termed Relational Chain-based Embedded KGQA. Novel techniques are proposed to effectively utilize QA relational chain parsing to identify the semantics more accurately and leverage the structure information preserved in KG embedding to reason the implicit answer indirectly.
Our comprehensive empirical results on three benchmarks demonstrate that our method outperforms many of its state-of-the-art counterparts. The experimental comparison between our approach and its ablated variants also verifies that the proposed model components contribute to the answer reasoning result.
We believe KGQA will continue to be an attractive and promising research direction with realistic industrial and domestic scenarios, such as Intelligent Recommendation, Smart Personal Assistant, Big Data Mining Services, and Automatic Customer Services.

In the future, we plan to study the following major problems: (i) To support real-world dynamic application scenarios, the KGQA application is always updated quickly and inevitably accumulates new and immense external knowledge in real time. How can we augment our available KG's knowledge reserve automatically and incrementally to expand our system's knowledge coverage? (ii) This model is trained on relatively small QA datasets under weak supervision without external prior knowledge. How can we introduce external knowledge such as knowledge from web pages and other open-domain KGs to improve our question answering system's performance?

Following the universal solution patterns of the KGQA task, the method presented in this paper assumes that ``Our model will always choose an optimal answer from the KG''. Therefore, the method based on this assumption is definitely not suitable for the case where the answer does not exist in the KG. In future work, we will add research on the sub-task of ``Detecting whether the answer exists in the KG''. Moreover, our proposed Rce-KGQA is essentially a pipeline mechanism, which could bring deviation propagation and poor performance. In future work, we will also enhance our \textit{Rce-KGQA} by integrating the separated \textit{Answer Filtering Module} and \textit{Relational Chain Reasoning Module} together. Concretely, the major factor, which hinders the joint modelling, is the multiple step-by-step answer retrieval due to the KG's traditional structured storage pattern. Inspired by Fabio et al. \cite{fabioemnlp} who prove that the huge-volume PLMs have surprising knowledge storing capabilities, we will try to infuse the KG knowledge into the Transformer-based PLMs, which can theoretically solve the end-to-end fashion modelling difficulty well from the root.

\section*{Acknowledgement}
This work was partially supported by the Shanghai Yangfan Program (Project Code: 22YF1413600), the Major Research Plan of National Natural Science Foundation of China (Project Code: 92167102), and the Shaanxi Province Key Industrial Chain Projects (Project Code: NO.2018ZDCXL-GY-04-03-02). The authors would like to thank Guizhong Liu and Ruiping Yin for providing helpful discussions and comments.

\bibliography{kbqa-references}


\end{document}